\pdfoutput=1

\documentclass[11pt]{article}

\usepackage[]{acl2023}

\usepackage{times}
\usepackage{latexsym}
\usepackage[normalem]{ulem}
\usepackage{booktabs}
\usepackage{amsmath}
\usepackage{amsfonts}
\usepackage{amsthm}
\usepackage{multirow}
\usepackage{multicol}
\usepackage{graphicx}
\usepackage{caption}
\usepackage[noabbrev]{cleveref}
\usepackage{subcaption}
\usepackage[T1]{fontenc}

\usepackage[utf8]{inputenc}

\usepackage{microtype}

\newtheorem{definition}{Definition}
\DeclareMathOperator*{\argmax}{argmax}

\newcommand{\nospacetext}[1]{\makebox[0pt][l]{#1}}

\newcommand{\maybe}[1]{\textcolor{gray}{}}

\newcommand{\trecb}[0]{\textsc{trec-2}}
\newcommand{\trec}[0]{\textsc{trec-6}}
\newcommand{\agnb}[0]{\textsc{agn-2}}
\newcommand{\agn}[0]{\textsc{agn-4}}
\newcommand{\subj}[0]{\textsc{subj}}

\newcommand{\bert}[0]{\textsc{bert}}
\newcommand{\electra}[0]{\textsc{electra}}

\newcommand{\beast}[0]{\textsc{beast}}
\newcommand{\alsbi}[0]{\textsc{alsbi}}

\newcommand{\auc}[0]{\textsc{auc}}
\newcommand{\lcr}[0]{\textsc{lcr}}

\newcommand{\st}[0]{\textsc{st}}
\newcommand{\et}[0]{\textsc{et}}
\newcommand{\etb}[0]{\textsc{et$^\mathcal{B}$}}
\newcommand{\etbns}[0]{\textsc{et\nospacetext{$^\mathcal{B}$}}}
\newcommand{\etar}[0]{\textsc{eta}}
\newcommand{\etab}[0]{\textsc{eta$^\mathcal{B}$}}
\newcommand{\etabns}[0]{\textsc{eta\nospacetext{$^\mathcal{B}$}}}

\newcommand{\rnd}[0]{\textsc{rnd}}
\newcommand{\ent}[0]{\textsc{ent}}
\newcommand{\rg}[0]{\textsc{rg}}
\newcommand{\cs}[0]{\textsc{cs}}
\newcommand{\dal}[0]{\textsc{dal}}
\newcommand{\mc}[0]{\textsc{mc}}

\title{Smooth Sailing: Improving Active Learning for Pre-trained Language Models with Representation Smoothness Analysis}

\author{Josip Juki{\'{c}} \quad Jan {\v{S}}najder\\
TakeLab\\
Faculty of Electrical Engineering and Computing, University of Zagreb, Croatia\\
\tt \{josip.jukic, jan.snajder\}@fer.hr
} 

\begin{document}
\maketitle

\begin{abstract}
Developed to alleviate prohibitive labeling costs, active learning (AL) methods aim to reduce label complexity in supervised learning. While recent work has demonstrated the benefit of using AL in combination with large pre-trained language models (PLMs), it has often overlooked the practical challenges that hinder the effectiveness of AL. We address these challenges by leveraging representation smoothness analysis to ensure AL is \textit{feasible}, that is, both effective and practicable. Firstly, we propose an early stopping technique that does not require a validation set -- often unavailable in realistic AL conditions -- and observe significant improvements over random sampling across multiple datasets and AL methods. Further, we find that task adaptation improves AL, whereas standard short fine-tuning in AL does not provide improvements over random sampling. Our work demonstrates the usefulness of representation smoothness analysis for AL and introduces an AL stopping criterion that reduces label complexity.\footnote{Our code is available at \url{https://github.com/josipjukic/al-playground}}
\end{abstract}

\section{Introduction}
\label{sec:intro}

The notorious data hungriness of deep learning models emphasizes the importance of efficient and effective label acquisition.
However, the labeling process is often tedious and expensive, ultimately slowing the development of labeled datasets and resulting in subpar models.
Evolved out of a practical necessity, \textit{active learning} \cite[\textbf{AL};][]{cohn-etal-1996-active, settles-2009-active} is a special family of machine learning algorithms designed to reduce \textit{label complexity} -- the number of labels that a learning algorithm requires to achieve a given performance \cite{dasgupta-2011-two} -- and thus minimize labeling costs.
An AL method aims to select the most informative examples, which can be particularly useful when unlabeled data are abundant, but the labeling is costly or requires substantial expertise.

The striking success of deep learning has motivated the use of traditional AL techniques for training deep neural networks (DNNs) and the development of novel AL methods suited specifically to DNNs.
In natural language processing (NLP), AL has been shown to outperform a random selection of examples in many NLP tasks \cite{zhang-etal-2017-active, siddhant-lipton-2018-deep, ikhwantri-etal-2018-multi}.
Before the widespread adoption of large pre-trained language models (PLMs), a typical AL approach to training deep models was to use task-specific neural models trained from scratch in each AL step \cite{kasai-etal-2019-low, prabhu-etal-2019-sampling}.
Since PLMs fine-tuned to downstream tasks outperform standard neural models, PLMs have supplanted most of them, and researchers have begun to investigate the feasibility of AL for PLMs \cite{ein-dor-etal-2020-active, schroder-etal-2022-revisiting}.
Recent work in AL experimented with several training regimes, such as PLM adaptation and specific fine-tuning techniques \cite{yuan-etal-2020-cold, margatina-etal-2022-importance}.
In particular, task-adaptive pre-training (\textbf{TAPT}) has emerged as a cost-effective method for performance improvement complementary to AL \cite{howard-ruder-2018-universal}. TAPT uses additional pre-training on the unlabeled training set via masked language modeling and self-supervision.
In theory, combining AL with adapted PLMs should produce greater reductions in label complexity than either of the methods in isolation. However, since research on combining AL with PLMs is still in its infancy, whether it can work consistently better than random selection in realistic conditions remains an open question.

One of the challenges in combining AL and PLMs is that, although AL is conceptually simple and promises efficiency gains, there are a host of practical challenges in deploying it in realistic conditions \cite{attenberg-provost-2011-inactive, lowell-etal-2019-practical}. The situation is further aggravated by the fact that most AL research overlooks these challenges and resorts to unrealistic evaluation setups and resources.
One of the most pervasive problems stems from using a hold-out set during training (e.g., a validation set for regularization by early stopping). In real applications, hold-out sets are unlikely to be available, as building them would require additional labeling effort the AL is meant to reduce in the first place.
Another major problem is the flawed evaluation of AL methods:  typically, an AL method is compared against random selection as the baseline, but the two training regimes are not kept identical, which confounds the measured effect of AL.
In addition to the above-mentioned problems, there is the important practical question of when to stop the acquisition of labels, i.e., how to define the AL \textit{stopping criterion}.

AL methods rely highly on the \textit{acquisition model} (the underlying model used for selecting examples). Therefore, it is important to maintain good generalization properties of the acquisition model, which can be analyzed using representation smoothness.
Recently, functional space theory has emerged as a valuable tool for analyzing generalization properties and expressivity of DNNs \cite{yarotsky-2017-error, suzuki-2019-adaptivity}.
In particular, the \textit{Besov space}, a general function space that can capture spatial inhomogeneity, appears convenient for such analyses \cite{suzuki-atsushi-2021-deep}.

In this work, we address the practical challenges of AL. First, we systematically evaluate the \textbf{feasibility}, where we consider an AL method to be feasible if it is both \textit{practicable} (achievable in realistic conditions) and \textit{effective} (consistently outperforms random selection). Concretely, we explore different learning regimes in AL on various NLP classification tasks without a validation set that is unavailable in most real-world labeling campaigns. Motivated by the effectiveness of TAPT for PLMs \citep{gururangan-etal-2020-dont}, we explore how TAPT combines with AL in the low-resource setup.
Secondly, we leverage the representation smoothness of PLM layers in the Besov space to improve AL effectiveness.
In particular, we develop \textit{Besov early stopping}, an early stopping regularization technique that does not require a validation set, and we show that it consistently improves the model performance and reduces the variance of results for all AL methods we consider. Moreover, Besov early stopping shows promise as a surrogate for a validation set in zero- and few-shot setups for regular training without AL.
We also utilize representation smoothness to develop a stopping criterion based on the smoothness of AL samples to minimize label complexity. Our experiments show a reduction in label complexity for PLMs across five NLP datasets and five AL methods.
In addition, building on the idea that representation smoothness is relevant for AL, we complement our experiments with a novel AL method based on the norm of representation gradients. Both the proposed method and the existing AL methods consistently outperform random selection on PLMs with TAPT, which supports the recent findings that the training regime is more important than the choice of the AL method \cite{margatina-etal-2022-importance}.

Our contributions can be summarized as follows:
(1) we conduct a systematic evaluation of AL methods for large PLMs and show that AL is feasible, i.e., it consistently outperforms random selection under realistic conditions,
(2) we analyze the smoothness of the representation space of PLMs in AL and propose an early stopping technique that improves AL performance and stabilizes the results,
(3) we discover patterns in the representation smoothness of AL samples, which we use for an effective AL stopping criterion, and
(4) we introduce a representation-based AL method, competitive with other state-of-the-art AL strategies.
Our results demonstrate that AL with PLMs is feasible. Even more importantly, the results indicate that representation smoothness analysis can be leveraged to improve model training in general and the effectiveness of AL in particular, opening new avenues for further research.

\section{Related Work}
\label{sec:rw}

Our work builds on several strands of research, including practical challenges in AL, combining AL with PLMs, and different training setups for AL acquisition models.

\paragraph{Practical challenges in AL.}
Despite the success of AL for many NLP tasks, studies have identified a number of practical challenges hindering the broader deployment of AL \cite{attenberg-provost-2011-inactive, lowell-etal-2019-practical}. The most obvious problem is the unavailability of a labeled validation set, an essential resource in model training typically used for hyperparameter optimization and regularization via early stopping. Moreover, in realistic AL conditions, a labeled test set is also unavailable, making a held-out evaluation of the underlying model's quality impossible. Previous work mostly used model confidence or training error stability to evaluate the acquisition model and derive an AL stopping criterion based on that estimation \cite{vlachos-2008-stopping, bloodgood-vijay-shanker-2009-method, zhu-etal-2010-confidence, ishibashi-hideitsu-2021-stopping}. However, these criteria have not been widely adopted as they often require tuning for specific datasets and tasks. We mitigate this by developing a task-agnostic AL stopping criterion that detects the points of the largest reduction in label complexity compared to random selection.

\paragraph{AL with PLMs.}
Only recently have large PLMs been coupled with AL. Early work concentrated mainly on the Transformer architecture \cite{vaswani-etal-2017-attention} utilizing a simple training setup. More concretely, the predominant approach was to use a standard fine-tuning technique with a fixed number of training epochs, fine-tuning the model from scratch in each AL step \cite{ein-dor-etal-2020-active, margatina-etal-2021-active, shelmanov-etal-2021-active, karamcheti-etal-2021-mind, schroder-etal-2022-revisiting}. However, \citet{mosbach-etal-2021-stability} and \citet{zhang-etal-2021-revisiting} showed that fine-tuning in low-resource setups (scenarios with little training data) tends to be very unstable, especially when training for only a few epochs. This instability poses a serious issue, as AL often implies a low-resource setting. Moreover, fine-tuning is often sensitive to weight initialization and data ordering \cite{dodge-etal-2020-fine}. This instability of PLM fine-tuning also makes the AL results unstable. We address the instability issue by proposing an early stopping technique without a validation set, and we show that combining PLMs with AL is feasible.

\paragraph{AL training regimes.}
AL research took a turn from standard fine-tuning of pre-trained models to explore different training regimes and how to use them in combination with AL methods. For example, \citet{griesshaber-etal-2020-fine} explored how to efficiently fine-tune Transformers with AL by freezing the network's layers. Similarly, \citet{yuan-etal-2020-cold} explored self-supervised language modeling to estimate example informativeness for cold-start active learning. Motivated by the general success of TAPT \cite{gururangan-etal-2020-dont}, \citet{margatina-etal-2022-importance} showed that AL outperformed random sampling for PLMs with TAPT, albeit using a validation set. Similarly, \citet{yu-etal-2022-actune} developed a self-training approach for active learning with the addition of weighted clustering. While some training regimes seem promising for AL, the outstanding question is which regimes can consistently outperform random selection. Furthermore, considering what resources are realistically available during training, the primary concern is whether we can apply these training regimes in realistic conditions.
\section{Representation in Besov Space}
\label{sec:besov}

Due to their remarkable flexibility and adaptivity, deep learning models have gained significant traction.
To explain these phenomena, researchers have leveraged function space theory to develop approximation and estimation error analysis \cite{yarotsky-2017-error, suzuki-2019-adaptivity}. Our work relies on a particular type of analysis based on the theory of Besov spaces.

\subsection{Besov space}
It has been shown that the expressive power of DNNs can be analyzed by specifying the target function's property such as \textbf{smoothness} \cite{petersen-voigtlander-2018-optimal, imaizumi-fukumizu-2019-deep}, i.e., the number of orders of continuous derivatives it has over some domain.
\textit{Besov space} has proven to be especially convenient for such analyses, as it allows spatially inhomogeneous smoothness with spikes and jumps, which we often encounter in high-dimensional deep learning.
In Besov spaces, the approximation error (expressivity)\footnote{The approximation error refers to the distance between the target function and the closest neural network function of a given architecture.} and estimation error (generalizability)\footnote{Estimation error refers to the distance between the ideal network function and an estimated network function.} depend on the properties of the representation space \cite{suzuki-atsushi-2021-deep}. Given these theoretical connections, representation space analysis can steer toward better generalization properties.

\subsection{Besov smoothness index}
\label{subsec:bsi}
We briefly describe the mathematical apparatus of the Besov space analysis, adopted with slight modifications from \cite{suzuki-2019-adaptivity, suzuki-atsushi-2021-deep}. Let $\Omega \in \mathbb{R}^d$ be a domain of functions. For a function $f: \Omega \to \mathbb{R}$ with a defined $p$-norm in $L_p$ (space of measurable functions with finite $p$-norm) and seminorm $|f|$ defined by $x \mapsto |f(x)|$, we define $\|f\|_{p} := \|f\|_{L^p(\Omega)} := (\int_{\Omega}{|f|^p dx})^{\frac{1}{p}}$ for $0 < p < \infty$. For $p = \infty$, we define $\|f\|_{\infty} := \|f\|_{L^{\infty}(\Omega)} := \sup_{x \in \Omega}{|f(x)|}$. 
\begin{definition}[Smoothness modulus]
For a function $f \in L^p(\Omega)$, $p \in (0, \infty]$, $t \in (0, \infty)$, $h \in \mathbb{R}^d$, and $r \in \mathbb{N}$, the $r$-th modulus of smoothness of $f$ is defined by
\begin{equation*}
    w_{r,p} (f, t) := \sup_{\|h\|_2 \leq t}{\|\Delta^r_h (f)\|_p},
\end{equation*}
where $\Delta^r_h (f)$ is the forward difference operator of the $r$-th order defined as $\Delta^r_h (f)(x) := \sum_{i=0}^r \binom{r}{i} (-1)^{r-i} f(x + ih)$ for $[x, x + rh] \in \Omega$, and $0$ otherwise.
\end{definition}
\begin{definition}[Besov space $(\mathcal{B}^{\alpha}_{p,q})$]
For $0 < p,q \leq \infty$, $\alpha > 0$, $r := \lfloor \alpha \rfloor + 1$, let the seminorm of the Besov space $\mathcal{B}^{\alpha}_{p,q}$ be
\begin{equation}
|f|_{\mathcal{B}^{\alpha}_{p,q}} := \left(\int_0^\infty \left(t^{-\alpha} w_{r,p}(f,t)\right)^q \frac{dt}{t} \right)^\frac{1}{q}
\label{eq:seminorm}
\end{equation}
for $q < \infty$. Let $|f|_{\mathcal{B}^{\alpha}_{p,q}} = \sup_{t>0}t^{-\alpha} w_{r,p}(f,t)$ for $q = \infty$. \textbf{Besov smoothness index} of $f$ is determined as the maximum index $\alpha$ for which the Besov seminorm is finite.
\end{definition}

Intuitively, the Besov smoothness index (\textit{Besov smoothness} for short) quantifies the properties of DNN's representation space. More specifically, a higher index indicates higher smoothness. Because the calculation of Besov smoothness (more precisely, the integral in \eqref{eq:seminorm}) is intractable, we have to rely on approximations. \citet{elisha-dekel-2016-wavelet, elisha-dekel-2017-function} proposed wavelet decomposition of a random forest (RF) for approximating Besov smoothness. Wavelet decomposition of the RF establishes an order of importance of the RF nodes, while RF uses the embedded representations of an arbitrary DNN as features. For classification problems, we can normalize the inputs to $[0,1]$ and transform the class labels into vectors in the $\mathbb{R}^{L-1}$ space by assigning each label to a vertex of a standard simplex, where $L$ is the number of classes. This gives us the $k$-th layer of a neural network as a function $f_k: [0,1]^d \to \mathbb{R}^{L-1}$. For a random forest consisting of $J$ estimators, \citet{elisha-dekel-2017-function} proceeded by approximating the errors of each estimator $\mathcal{T}_j$ with $M$ most important wavelets. The error function (with $r=1$, $p=2$) is estimated as $\sigma_M \sim c_kM^{-\alpha_k}$. Numerically, we can use an approximation $\log(\sigma_m) \sim \log(c_k) - \alpha \log(m),$ $m=1,\dots,M$, and find $c_k$ and $\alpha_k$ through least squares, where $\alpha_k$ is the estimate for the Besov smoothness of $f_k$, i.e., the $k$-th layer of a DNN.

\subsection{Representation smoothness}
Analyzing the Besov smoothness of DNNs can unveil their representation geometry. DNNs should benefit from smoother representations, as they help the model avoid overfitting. Intuitively, ``well-learned'' representations will exhibit high Besov smoothness. When we decompose a PLM into wavelets sorted by relevance and use the Besov smoothness approximation described in \Cref{subsec:bsi}, smoother representations achieve lower generalization errors with fewer wavelets.

Another relevant phenomenon for representation smoothness analysis is that the individual layers of DNNs specialize in different features. In particular, earlier layers tend to learn \textit{generalization} features, while the deeper layers are more prone to \textit{memorization} \cite{stephenson-etal-2021-geometry, baldock-etal-2021-deep}. Following these insights, we propose using Besov smoothness to inspect the generalization properties of PLMs through the prism of layer-wise representation geometry. We hypothesize PLMs should benefit more from smoother representations in earlier layers, and we propose methods to enforce learning such representations during training.
\section{Preliminaries}
\label{sec:prelims}

In this section, we describe our experimental setup, detailing the datasets, models, AL methods, and evaluation metrics.

\subsection{Datasets}

We select three different single-text classification tasks commonly used in the AL literature. The datasets vary in size, number of classes, and complexity, allowing for a nuanced study of AL methods. To extend our analysis to similar datasets with different levels of complexity, we also add binary versions of the multi-class tasks. In total, we work with five datasets (cf.~Appendix, \Cref{tab:dataset-stats}): (1) the question type classification dataset \cite[\textbf{\trec{}};][]{li-roth-2002-learning}; (2) the corresponding binary version \textbf{\trecb{}} with only the two most frequent classes (\textit{Entity} and \textit{Human}); (3) the subjectivity dataset\textbf{ \subj{}} of \citet{pang-lee-2004-sentimental}, which classifies the movie snippets as subjective or objective and is often used in AL benchmarks; (4) the AG's News classification dataset \textbf{\agn{}} of \citet{zhang-etal-2015-character}; and (5) its binary version, \textbf{\agnb{}}, often used in the AL literature, with two categories (\textit{World} and \textit{Sports}) out of four.

\subsection{Models}
We focus on large PLMs and include two representatives of the Transformer family, each using a different pre-training paradigm. Specifically, we experiment with \bert{} \cite{devlin-etal-2019-bert}, which uses a generative pre-training approach via masked language modeling, and \electra{}  \cite{clark-etal-2020-electra}, which relies on discriminative training to detect corrupted tokens induced by a small generator network.
For both models, we leverage their widely used \textit{base} variants from the Hugging Face library \cite{wolf-etal-2020-transformers}, which consist of $12$ layers.

\subsection{Active learning methods}
We consider six sampling strategies, including random selection, which serves as a baseline. The other five strategies are AL methods from different families.

\begin{description}

\item[\textbf{Random selection}] (\textbf{\rnd{}}) selects instances uniformly from the unlabeled pool.
    
\item[\textbf{Maximum entropy}] \cite[\textbf{\ent{}};][]{lewis-gale-1994-sequential} comes from the family of \textit{uncertainty} strategies. The method queries instances where the model is least certain, according to the criterion of maximum entropy of the prediction output.

\item[\textbf{Monte Carlo dropout}]  \cite[\textbf{\mc{}};][]{gal-ghahramani-2016-dropout} is similar to \textsc{entropy}, but relies on the stochasticity of forward passes with dropout layers \cite{srivastava-etal-dropout} to estimate the entropy for a given instance.

\item[\textbf{Core-set}] \cite[\textbf{\cs{}};][]{sener-savarese-2018-active} promotes instance diversity by leveraging the learned representations of the acquisition model. The method aims to minimize the distance between an example in the unlabeled set and its most similar counterpart in the labeled subset.

\item[\textbf{Discriminative active learning}] \cite[\textbf{\dal{}};][]{gissin-shwartz-2019-discriminative} frames active learning as a classification of whether a particular instance is labeled or not to make the labeled and unlabeled sets indistinguishable. Specifically, \dal{} queries instances that are most likely to be in the unlabeled subset according to a trained classifier.

\item[\textbf{Representation gradients}] (\textbf{\rg{}}) is a novel AL strategy we propose in this work. Similar to methods from \cite{huang-etal-2016-active, ash-etal-2019-deep}, \rg{} selects instances based on gradient information from the representation space. However, unlike other gradient-based methods, \rg{} is much less computationally demanding and, therefore, suitable for resource-limited studies and realistic conditions. The method computes the mean representation gradient with respect to the embedded inputs and selects the instances with the largest gradient norm. Formally, with $\bar{\mathbf{h}}$ as the mean representation, the \rg{}'s selection criterion is
$
    \argmax_{\mathbf{x} \in \mathcal{U}}{
    \left\|
    \partial_{\mathbf{x}}\bar{\mathbf{h}}
    \right\|_2}
    ,
$

where $\mathcal{U}$ denotes the unlabeled set. The intuition behind \rg{} is that the locally sharp instances in the representation space of the underlying model, i.e., the ones with large gradient norms, surprise the model the most and thus will contribute the most to a reduction in label complexity.

\end{description}

In our experiments, we select $50$ new examples in each step of each AL experiment, using $100$ examples for the warm start (randomly sampled labeled data to kick-start the model). We set the labeling budget to $1,000$ instances for easier datasets (\trecb{}, \agnb{}, and \subj{}) and $2,000$ instances for harder datasets (\trec{} and \agn{}). 


\subsection{Evaluation}
To evaluate the entire AL process, we use the area under the performance curve (\textbf{\auc{}}). Each step corresponds to the classification performance in terms of the $F_1$ score of a model trained with a certain number of labeled examples. We advocate using \auc{} complementary to the AL curves, as we believe it is a good approximation of AL feasibility as a summary numeric score. Since we use different training regimes in our experiments, we compare each AL strategy to random selection within the same training regime to isolate the effects of AL.
Additionally, we introduce a metric to measure the direct practical gains of AL by estimating the reduction in label complexity of AL compared to random selection. For a given AL step, we compute the number of additional labels required to achieve the same performance with random selection, thus estimating the number of labels one saves when using AL. We refer to this metric as \textit{label complexity reduction} (\textbf{\lcr{}}). 
\section{Improving Active Learning}
\label{sec:beast}

In this section, we first look into the representation geometry of PLMs by means of representation smoothness analysis. Then, we link our findings to devise a smoothness-based early stopping technique that does not require a validation set. We explore the effects of our method in different training regimes and provide a systematic evaluation of AL for PLMs in the low-resource setup.

\subsection{Representation smoothness analysis}

We empirically test the characteristics of Besov smoothness of PLMs. In particular, we compare the representation smoothness of PLMs in three different training regimes: (1) \textbf{short training} (\textsc{st}), where models were trained for $5$ epochs, (2) \textbf{extended training} (\textsc{et}), where models were trained for $15$ epochs, and (3) model adaptation with TAPT (cf.~\Cref{app:hyper} for details) followed by an extended training for $15$ epochs (\textsc{eta}). We computed the smoothness of PLM layers during training, averaged across AL steps. In each AL step, we fine-tuned the model anew.

\begin{figure}[t]
\centering
\includegraphics[width=\linewidth]{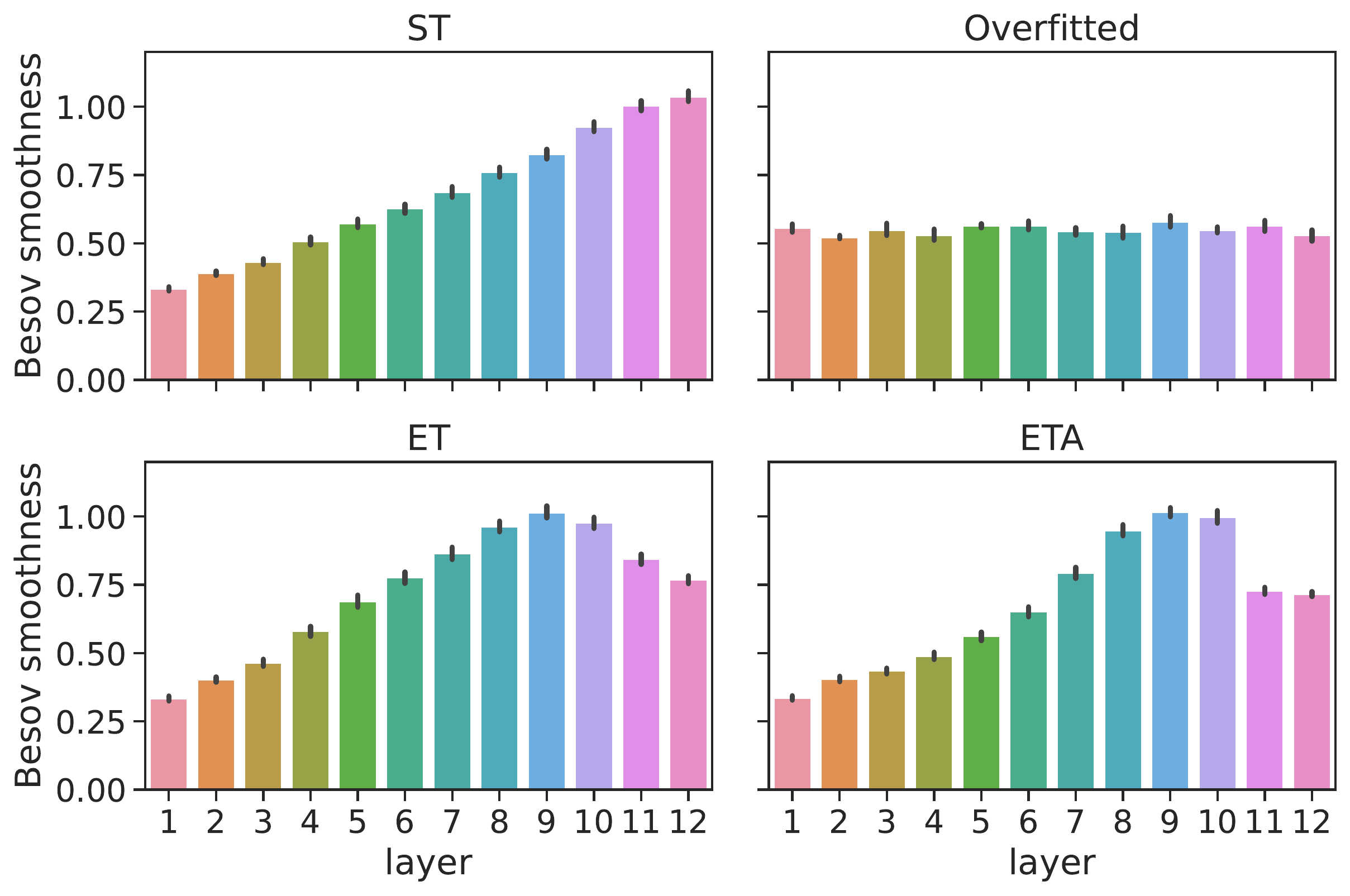}
\caption{Besov smoothness of PLM layers for different training regimes. The scores are normalized (between $0$ and $1$ per layer) and averaged across datasets, models, and AL methods. The black error bars represent the standard deviation. We note that the deviation is small, indicating similar behavior across different datasets, models, and AL methods.}
\label{fig:besov-layers}
\end{figure}

Performance-wise, \textsc{eta} yields better results than \et{} and \st{} (\Cref{tab:auc}). Moreover, AL in the \textsc{st} regime does not yield improvement over random sampling. \Cref{fig:besov-layers} shows the layer-wise smoothness for the three mentioned regimes with the addition of the overfitting regime, where we purposefully overfitted the acquisition model in each AL step by training the model for $100$ epochs. In the \st{} regime, we observe a monotonic increase in smoothness as we progress through layers, while the smoothness in \et{} peaks before the last few layers. The shift of the smoothness peak is even more pronounced for TAPT with extended training. In overfitted models, we observe a flat distribution of smoothness across layers. We observe that better performance and effective AL come with a shift in smoothness distribution towards earlier layers, as displayed in \et{} and \etar{} regimes. We hypothesize that, in the low-resource setup, the deeper layers exhibit higher smoothness in the \st{} regime because they are prone to \textit{heuristic memorization} -- DNN relies on spurious artifacts (shortcuts) that are correlated with a target label \cite{bansal-etal-2022-measures} -- which may cause the model to perform poorly.

\subsection{Besov early stopping}

In the AL loop, the effect of selecting an acquisition model with poor generalization properties propagates through the AL steps.
To ensure the effectiveness of AL, regularization by early stopping is often used to pre-empt overfitting in order to retain good generalization properties. However, since a validation set is often unavailable in realistic conditions, using it for early stopping renders AL impracticable. However, feasible AL needs to be both effective and practicable.

The above empirical findings on smoothness distribution across the PLM layers for the different training regimes motivate an early stopping heuristic based on representation smoothness without a validation set. We propose \textbf{\beast{}} (\textbf{B}esov \textbf{ea}rly \textbf{st}opping), where we proceed with the training as long as the Besov smoothness distribution skews toward earlier layers. We define the stopping point as the epoch where the distribution skewness\footnote{We compute the layer-wise smoothness skewness as the Fisher-Pearson coefficient of skewness.} fails to increase, i.e., when the peak of the representation smoothness ([\textsc{cls}] token) fails to shift towards earlier layers for two consecutive epochs. We revert the model to the last epoch where this effect is preserved. In this way, we stop the training before the smoothness distribution flattens out, which we observe in overfitted models. We experiment with two more training regimes: \etb{} and \etab{}, which are just \et{} and \etar{} with \beast{}.

\begin{figure}[t]
\centering
\includegraphics[width=\linewidth]{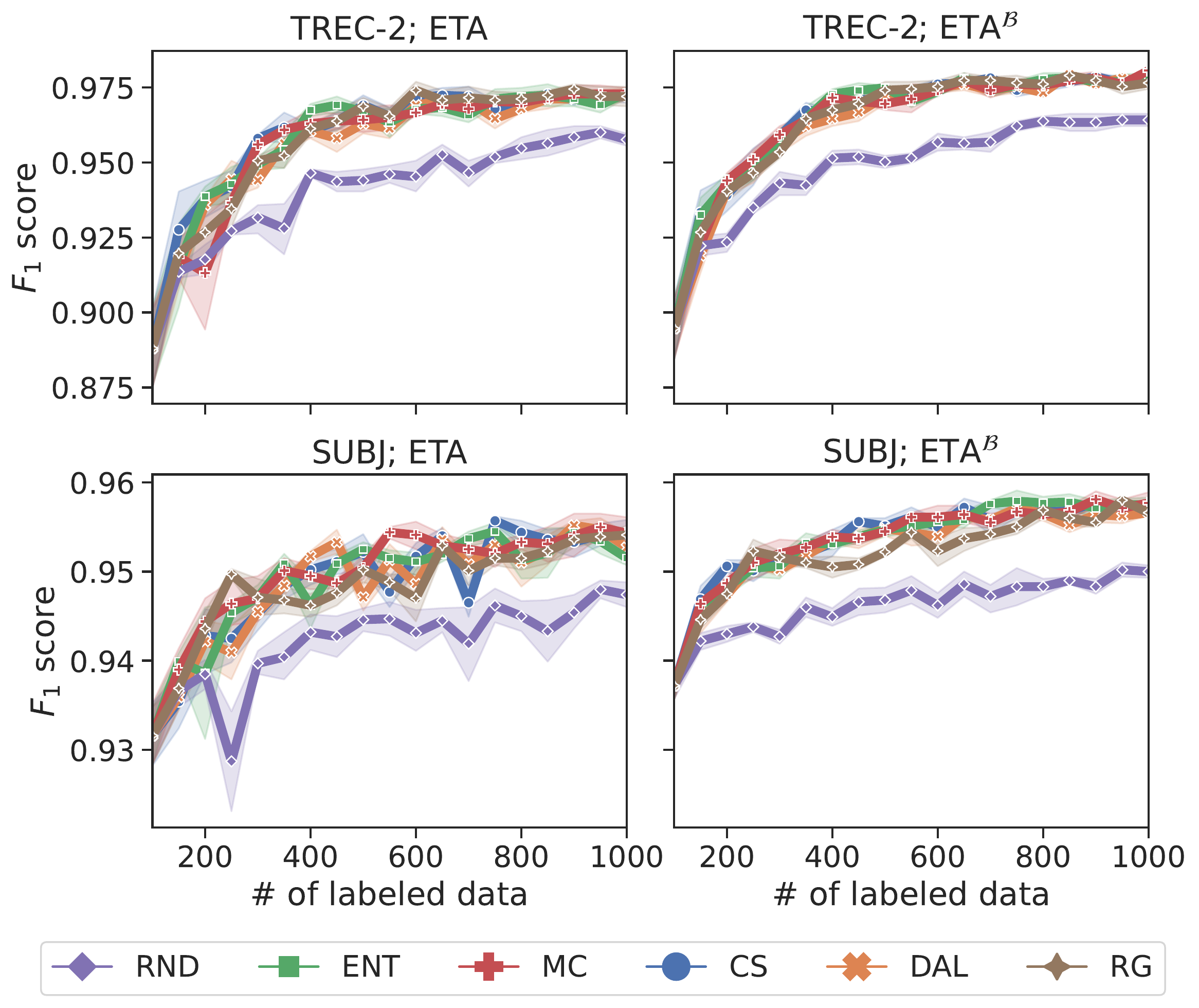}
\caption{Active learning performance curves for \bert{} in terms of $F_1$ score. Random sampling (purple rhombs) serves as a baseline. For the sake of space, we show the results on a subset of datasets for \bert{} and regimes \textsc{eta} and \textsc{eta$^\mathcal{B}$} as we obtained similar results for other configurations (cf.~\Cref{fig:app-curves} in Appendix). The results are averaged over five runs. The confidence intervals represent the standard deviation. Best viewed on a computer screen.}
\label{fig:al-plots}
\end{figure}

We compare \beast{} to the approaches without early stopping, where we chose the models from the last epoch. Our experiments show the difference in AL performance across different training regimes. \Cref{fig:al-plots} shows the trend of AL curves through the steps, and \Cref{tab:auc} provides more comprehensive comparisons with \auc{} as the aggregated measure of AL effectiveness.
We can observe that AL coupled with \textsc{st} performs poorly, and AL fails to outperform random sampling (sometimes even worse than random sampling).
The \et{} regime generally improves performance, with AL sometimes outperforming random selection.
\etar{} and \etab{} further improve performance over random sampling for every AL method on every dataset we used. For \bert{}, the difference between AL and random sampling is statistically significant in $22$ out of $25$ cases with \etar{} and in all $25$ cases with \etab{}.
More importantly, \etb{} and \etab{} outperform their counterparts without \beast{} and reduce the variance of the results (cf.~Appendix, \Cref{tab:std}).
We support the hypothesis that the choice of the AL method is not as important as the training regime, as we achieve similar results for every method when AL outperforms random selection. TAPT works across the board, improving AL performance on all five datasets. With the addition \beast{}, we achieve feasible AL, making it both practicable and effective. On top of that, even with random sampling, \beast{} consistently yields higher scores than the model from the last epoch, showing benefits even for regular fine-tuning without AL.

\begin{table}[t]
\centering
\small
\begin{tabular}{lrrrrrrr}
\toprule
& & \rnd{} & \ent{} & \mc{} & \cs{} & \dal{} & \rg{} \\
\midrule
\multirow{5}{*}{\rotatebox[origin=c]{90}{\trecb{}}}
& \st{} & $.875$ & $.873$ & $.883$ & $.881$ & $.889$ & $.879$ \\
& \et{} & $.912$ & $.932$\nospacetext{$^\dagger$} & $.934$\nospacetext{$^\dagger$} & $.929$ & $.931$ & $.931$\\
& \etbns{} & $.925$ & $.942$\nospacetext{$^\dagger$} & $.942$\nospacetext{$^\dagger$} & $.939$\nospacetext{$^\dagger$} & $.940$\nospacetext{$^\dagger$} & $.938$\\
& \etar{} & $.941$ & $.959$\nospacetext{$^\dagger$} & $.957$\nospacetext{$^\dagger$} & $.960$\nospacetext{$^\dagger$} & $.957$\nospacetext{$^\dagger$} & $.958$\nospacetext{$^\dagger$}\\
& \etabns{}
& $.949$ & $\mathbf{.966}$\nospacetext{$^\dagger$} & $.965$\nospacetext{$^\dagger$} & $.965$\nospacetext{$^\dagger$} & $.964$\nospacetext{$^\dagger$} & $.965$\nospacetext{$^\dagger$}\\
\midrule
\multirow{5}{*}{\rotatebox[origin=c]{90}{\subj{}}}
& \st{} & $.896$ & $.892$ & $.885$ & $.901$ & $.898$ & $.892$\\
& \et{} & $.920$ & $.922$ & $.922$ & $.925$ & $.925$ & $.920$\\
& \etbns{} & $.928$ & $.931$ & $.932$ & $.932$ & $.933$ & $.930$\\
& \etar{} & $.942$ & $.949$\nospacetext{$^\dagger$} & $.950$\nospacetext{$^\dagger$} & $.949$\nospacetext{$^\dagger$} & $.949$\nospacetext{$^\dagger$} & $.948$\\
& \etabns{} & $.946$ & $\mathbf{.954}$\nospacetext{$^\dagger$} & $\mathbf{.954}$\nospacetext{$^\dagger$} & $\mathbf{954}$\nospacetext{$^\dagger$} & $.953$\nospacetext{$^\dagger$} & $.952$\nospacetext{$^\dagger$}\\
\midrule
\multirow{5}{*}{\rotatebox[origin=c]{90}{\agnb{}}}
& \st{} & $.923$ & $.942$\nospacetext{$^\dagger$} & $.941$\nospacetext{$^\dagger$} & $.922$ & $.941$\nospacetext{$^\dagger$} & $.942$\nospacetext{$^\dagger$}\\
& \et{} & $.960$ & $.969$ & $.970$\nospacetext{$^\dagger$} & $.965$ & $.967$ & $.969$\\
& \etbns{} & $.967$ & $.974$\nospacetext{$^\dagger$} & $.975$\nospacetext{$^\dagger$} & $.972$ & $.974$\nospacetext{$^\dagger$} & $.975$\nospacetext{$^\dagger$}\\
& \etar{} & $.974$ & $.981$\nospacetext{$^\dagger$} & $.980$ & $.981$\nospacetext{$^\dagger$} & $.980$ & $.980$\nospacetext{$^\dagger$}\\
& \etabns{} & $.977$ & $\mathbf{.983}$\nospacetext{$^\dagger$} & $\mathbf{.983}$\nospacetext{$^\dagger$} & $\mathbf{.983}$\nospacetext{$^\dagger$} & $.982$\nospacetext{$^\dagger$} & $.982$\nospacetext{$^\dagger$}\\
\midrule
\multirow{5}{*}{\rotatebox[origin=c]{90}{\trec{}}}
& \st{} & $.706$ & $.743$\nospacetext{$^\dagger$} & $.749$\nospacetext{$^\dagger$} & $.666$ & $.689$ & $.693$\\
& \et{} & $.867$ & $.878$ & $.881$\nospacetext{$^\dagger$} & $.867$ & $.878$ & $.867$\\
& \etbns{} & $.873$ & $.885$ & $.890$\nospacetext{$^\dagger$} & $.873$ & $.882$ & $.875$\\
& \etar{} & $.909$ & $.933$\nospacetext{$^\dagger$} & $.931$\nospacetext{$^\dagger$} & $.931$\nospacetext{$^\dagger$} & $.934$\nospacetext{$^\dagger$} & $.930$\nospacetext{$^\dagger$}\\
& \etabns{} & $.925$ & $.939$\nospacetext{$^\dagger$} & $.937$\nospacetext{$^\dagger$} & $.936$\nospacetext{$^\dagger$} & $\mathbf{.940}$\nospacetext{$^\dagger$} & $.935$\nospacetext{$^\dagger$}\\
\midrule
\multirow{5}{*}{\rotatebox[origin=c]{90}{\agn{}}}
& \st{} & $.837$ & $.828$ & $.824$ & $.801$ & $.834$ & $.829$\\
& \et{} & $.869$ & $.869$ & $.871$ & $.871$ & $.880$\nospacetext{$^\dagger$} & $.875$\\
& \etbns{} & $.875$ & $.877$ & $.878$ & $.879$ & $.886$\nospacetext{$^\dagger$} & $.881$\\
& \etar{} & $.891$ & $.905$\nospacetext{$^\dagger$} & $.905$\nospacetext{$^\dagger$} & $.902$\nospacetext{$^\dagger$} & $.906$\nospacetext{$^\dagger$} & $.899$\nospacetext{$^\dagger$}\\
& \etabns{} & $.894$\nospacetext{$^\dagger$} & $.908$\nospacetext{$^\dagger$} & $.908$\nospacetext{$^\dagger$} & $.905$\nospacetext{$^\dagger$} & $\mathbf{.909}$\nospacetext{$^\dagger$} & $.903$\nospacetext{$^\dagger$}\\
\bottomrule
\end{tabular}
\caption{\auc{} scores for random sampling and different AL methods across datasets and training regimes for \bert{} (cf.~Appendix, \Cref{tab:app-auc}). The results are averaged over $5$ runs with different seeds. \textbf{Bold} numbers indicate the best \auc{} for each dataset. The ``$\dagger$'' indicates when the mean \auc{} of an AL method is significantly different from random sampling (two-sided Man-Whitney U test with $p<.05$, adjusted for family-wise error rate with the Holm-Bonferroni method).}
\label{tab:auc}
\end{table}

\section{Active Sample Smoothness}
In \Cref{sec:beast}, we analyzed the Besov smoothness of layer representations of PLMs. In this section, we take a step further and examine the smoothness at the instance level. Instead of using the representations on the training set, we computed the Besov smoothness as the average across layers on the unseen (selected but not yet trained on) \textit{active sample}, acquired by the AL method. In contrast to the seen training examples, we argue that the Besov smoothness on unseen examples can be interpreted as the amount of information the model could obtain from that sample. More precisely, the lower the smoothness of an active sample, the more informative it is for the model. In contrast, smooth samples are already well-represented and thus not as resource-effective as their less smooth counterparts.

\begin{figure}[t!]
\small
\centering
\begin{subfigure}{.49\linewidth}
  \centering
  \includegraphics[width=\linewidth]{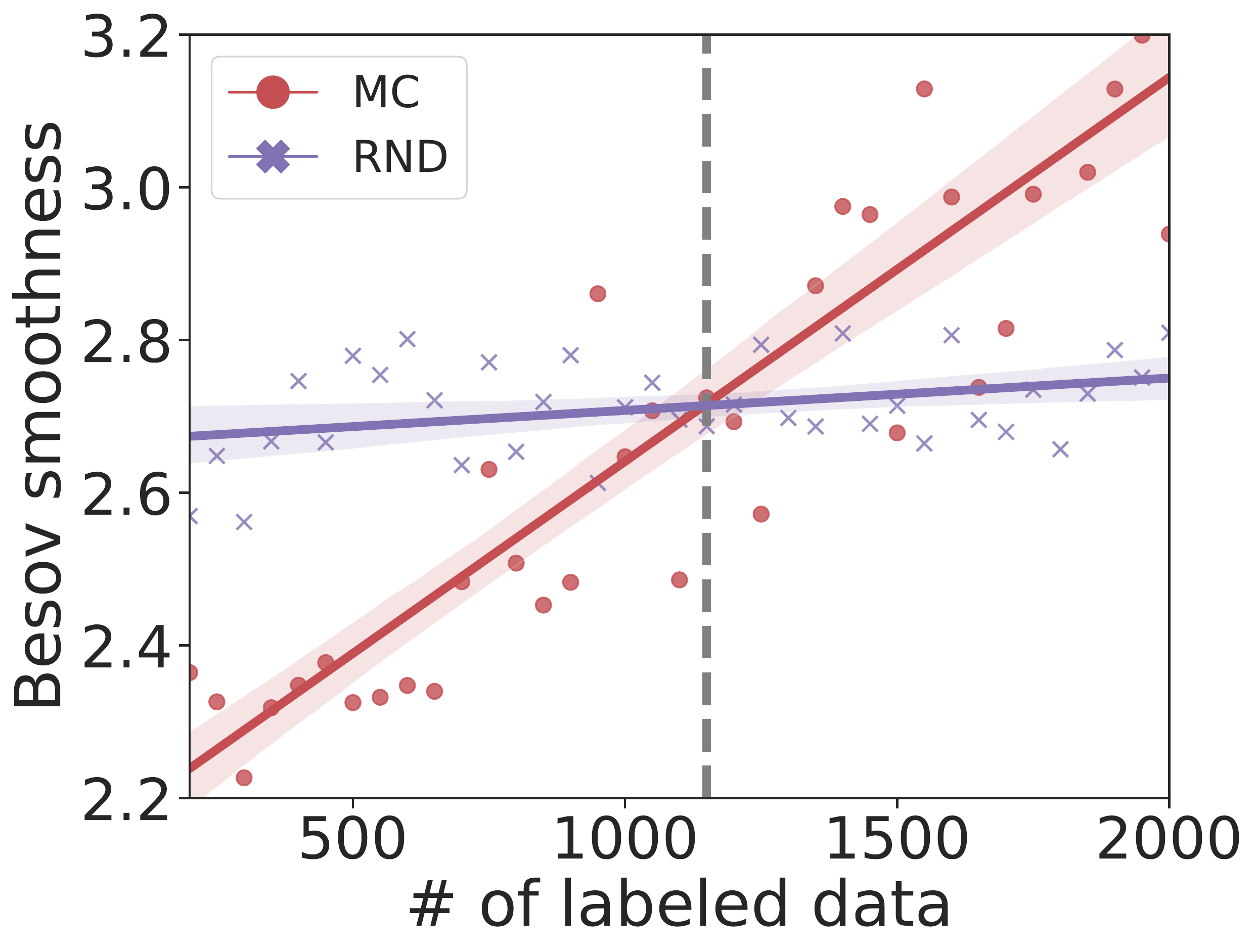}
  \caption{Sample smoothness}
  \label{fig:al-stop-smooth}
\end{subfigure}
\begin{subfigure}{.49\linewidth}
  \centering
  \includegraphics[width=\linewidth]{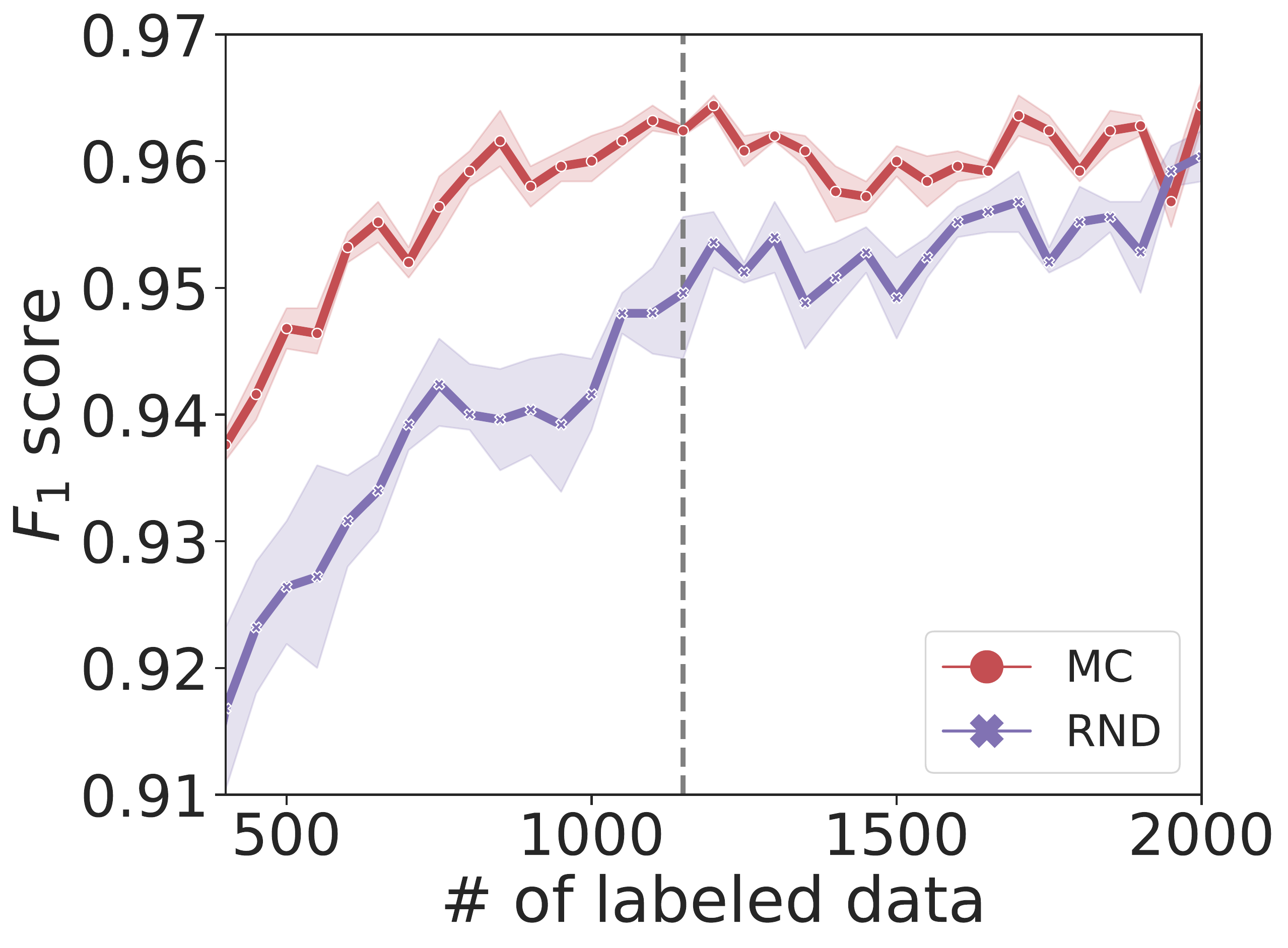}
  \caption{AL curve}
  \label{fig:al-stop-curve}
\end{subfigure}
\caption{The relationship between an active sample and random sample smoothness. \Cref{fig:al-stop-smooth} shows how the smoothness of samples retrieved by AL (red) relates to the smoothness of random samples (violet) with fitted regression lines. The smoothness values are calculated as the average across layers. \Cref{fig:al-stop-curve} shows the corresponding AL performance curve. The gray dotted line indicates the intersection of active and random sample smoothness, which signals the beginning of diminishing returns of AL. We show the results for \textsc{bert} in the \etab{} regime on the \textsc{trec-6} dataset as an illustration. We observe very similar patterns in other datasets. Since all of the AL methods display similar behavior in \etab{} regime, we show only \mc{} to avoid clutter (cf.~\Cref{fig:app-active-sample} in Appendix for other datasets). Best viewed on a computer screen.}
\label{fig:active-sample}
\end{figure}

We compare the Besov smoothness of actively acquired samples against random samples. We consistently observe two patterns, showcased by \Cref{fig:active-sample}. First, the smoothness of random samples is uniform throughout the AL steps. The second pattern occurs in the trend of AL sample smoothness. In the early AL steps, AL sample smoothness is low, indicating sharp representations that require smoothing (by learning). As the AL procedure progresses, the acquisition model improves, and the active samples' smoothness increases. We interpret this as the model slowly consuming the information from the data pool, eventually reaching a state of ``information depletion'', i.e., a state in which the remaining unlabeled data provides no additional value to the model.

\begin{table}[tb!]
\small
\centering
\begin{tabular}{lrrrrrr}
\toprule
& & \textsc{ent} & \textsc{mc}  & \textsc{cs} & \textsc{dal}  & \textsc{rg}\\
\midrule
\multirow{2}{*}{\rotatebox[origin=c]{0}{\etar{}}}
& avg & $.316$ & $.312$ & $.351$ & $.275$ & $.319$\\
& \alsbi{} & $.447$ & $.401$ & $.511$ & $.282$ & $\textbf{.521}$\\
\midrule
\multirow{2}{*}{\rotatebox[origin=c]{0}{\textsc{eta\nospacetext{$^\mathcal{B}$}}}}
& avg & $.385$ & $.385$  & $.394$ & $.354$ & $.382$ \\
& \alsbi{} & $.586$ & $.532$ & $.488$ & $.447$ & $\textbf{.645}$\\
\bottomrule
\end{tabular}
\caption{Average \textsc{lcr} across datasets and models. The scores indicate the proportion of the dataset that needs to be labeled for random sampling to match the performance of the corresponding AL method. \alsbi{} is compared to an average \lcr{} throughout the AL steps (avg). The results are averaged over $5$ runs. Numbers in \textbf{bold} indicate the largest \textsc{lcr} for a certain training regime.}
\label{tab:label-complexity}
\end{table}

\paragraph{Stopping criterion.}
Our preliminary analysis of the relationship between the active and random sample smoothness motivates a simple stopping criterion, which we refer to as \alsbi{} (\textbf{a}ctive \textbf{l}earning \textbf{s}topping by \textbf{B}esov \textbf{i}ndex). \alsbi{} aims to detect when AL methods reach information depletion. We terminate the AL process when sample smoothness surpasses the average smoothness of a random sample in two consecutive steps. We disregard the first AL step, as it often takes several steps for the acquisition models to stabilize. Since we cannot compute the smoothness of a random sample (as we query only AL samples) in realistic conditions, we estimate the random sample smoothness on the warm start examples via bootstrapping. This approximation proved stable for $100$ examples as the smoothness of a random sample remains stable throughout AL steps. We take the average smoothness of $1,000$ bootstrapped samples of size $50$. \Cref{tab:label-complexity} shows that \alsbi{} yields larger \lcr{} than what one would get on average across AL steps, which supports our preliminary analysis. \textsc{rg} achieves the highest \lcr{} among the tested AL methods, which we believe is due to its compatibleness to \alsbi{} as both the AL method and the stopping criterion are based on representation smoothness. 

\section{Conclusion}
In our paper, we leverage representation smoothness analysis to improve the effectiveness of active learning (AL). In realistic conditions, we show that AL with pre-trained language models (PLMs) is effective when combined with task adaptation, while standard short fine-tuning often fails. We address the problem of unavailable resources (labeled hold-out sets) by developing the \textbf{B}esov \textbf{ea}rly \textbf{st}opping technique (\beast{}) that does not require a validation set. For AL to be feasible, it must be both effective and practicable. \beast{} meets both feasibility requirements: it improves AL performance over random sampling and reduces the variance of the performance scores across AL steps (effectiveness) while not requiring additional labeled data (practicability). Moreover, \beast{} improves the performance of PLMs even in standard fine-tuning without AL, which makes it potentially useful in zero-shot and few-shot setups where a validation set could also be unavailable. We further show the usefulness of representation smoothness analysis for AL by devising a simple and effective AL stopping criterion. We corroborate the hypothesis from previous research in that the effectiveness of AL is influenced more by the training regime rather than the AL method itself. We believe that the relationship between PLMs' generalization properties, label complexity, and representation smoothness is an exciting avenue for AL, and we hope our results will motivate further research in that direction.

\section*{Limitations}
\label{sec:limitations}

To fully comprehend the significance of our findings, it is necessary to consider the limitations of this study. Firstly, we evaluate only two Transformer-based models on a small number of text classification tasks. Although we used the models with different pre-training paradigms, it is possible that the findings do not generalize across models within the same family. In addition, we used the base variants of \textsc{bert} and \textsc{electra}, which both feature $12$ layers. Since our early stopping criterion is influenced by the number of layers whose smoothness we approximate, there is a possibility that smoothness would distribute differently for models with more or fewer layers. Another limitation is that we did not investigate these models' performance on tasks other than text classification, and the results may not be generalizable to different types of NLP tasks.

Since there are many different ways to measure the quality of an AL stopping criterion and we only wanted to illustrate the usefulness of smoothness patterns, we only compared the proposed \textsc{alsbi} method against an average baseline. However, a more comprehensive comparison with other approaches from the literature would provide a better understanding of the merit of our method.

 Lastly, we only scratched the surface of different training regimes for PLMs in the context of AL. Many new training regimes are emerging in the field, especially the ones focused on efficiency and modularity. We leave the exploration of these methods for future work.

\bibliography{references/anthology, references/custom}

\begin{thebibliography}{52}
\expandafter\ifx\csname natexlab\endcsname\relax\def\natexlab#1{#1}\fi

\bibitem[{Ash et~al.(2019)Ash, Zhang, Krishnamurthy, Langford, and
  Agarwal}]{ash-etal-2019-deep}
Jordan~T. Ash, Chicheng Zhang, Akshay Krishnamurthy, John Langford, and Alekh
  Agarwal. 2019.
\newblock \href {http://arxiv.org/abs/1906.03671} {Deep batch active learning
  by diverse, uncertain gradient lower bounds}.
\newblock \emph{CoRR}, abs/1906.03671.

\bibitem[{Attenberg and Provost(2011)}]{attenberg-provost-2011-inactive}
Josh Attenberg and Foster Provost. 2011.
\newblock Inactive learning? {D}ifficulties employing active learning in
  practice.
\newblock \emph{ACM SIGKDD Explorations Newsletter}, 12(2):36--41.

\bibitem[{Baldock et~al.(2021)Baldock, Maennel, and
  Neyshabur}]{baldock-etal-2021-deep}
Robert Baldock, Hartmut Maennel, and Behnam Neyshabur. 2021.
\newblock \href
  {https://proceedings.neurips.cc/paper/2021/file/5a4b25aaed25c2ee1b74de72dc03c14e-Paper.pdf}
  {Deep learning through the lens of example difficulty}.
\newblock In \emph{Advances in Neural Information Processing Systems},
  volume~34, pages 10876--10889. Curran Associates, Inc.

\bibitem[{Bansal et~al.(2022)Bansal, Pruthi, and
  Belinkov}]{bansal-etal-2022-measures}
Rachit Bansal, Danish Pruthi, and Yonatan Belinkov. 2022.
\newblock \href {https://openreview.net/forum?id=CCahlgHoQG} {Measures of
  information reflect memorization patterns}.
\newblock In \emph{Advances in Neural Information Processing Systems}.

\bibitem[{Bloodgood and
  Vijay-Shanker(2009)}]{bloodgood-vijay-shanker-2009-method}
Michael Bloodgood and K.~Vijay-Shanker. 2009.
\newblock \href {https://aclanthology.org/W09-1107} {A method for stopping
  active learning based on stabilizing predictions and the need for
  user-adjustable stopping}.
\newblock In \emph{Proceedings of the Thirteenth Conference on Computational
  Natural Language Learning ({C}o{NLL}-2009)}, pages 39--47, Boulder, Colorado.
  Association for Computational Linguistics.

\bibitem[{Clark et~al.(2020)Clark, Luong, Le, and
  Manning}]{clark-etal-2020-electra}
Kevin Clark, Minh-Thang Luong, Quoc~V. Le, and Christopher~D. Manning. 2020.
\newblock \href {https://openreview.net/pdf?id=r1xMH1BtvB} {{ELECTRA}:
  Pre-training text encoders as discriminators rather than generators}.
\newblock In \emph{ICLR}.

\bibitem[{Cohn et~al.(1996)Cohn, Ghahramani, and
  Jordan}]{cohn-etal-1996-active}
David~A Cohn, Zoubin Ghahramani, and Michael~I Jordan. 1996.
\newblock Active learning with statistical models.
\newblock \emph{Journal of artificial intelligence research}, 4:129--145.

\bibitem[{Dasgupta(2011)}]{dasgupta-2011-two}
Sanjoy Dasgupta. 2011.
\newblock \href {https://doi.org/https://doi.org/10.1016/j.tcs.2010.12.054}
  {Two faces of active learning}.
\newblock \emph{Theoretical Computer Science}, 412(19):1767--1781.
\newblock Algorithmic Learning Theory (ALT 2009).

\bibitem[{Devlin et~al.(2019)Devlin, Chang, Lee, and
  Toutanova}]{devlin-etal-2019-bert}
Jacob Devlin, Ming-Wei Chang, Kenton Lee, and Kristina Toutanova. 2019.
\newblock \href {https://doi.org/10.18653/v1/N19-1423} {{BERT}: Pre-training of
  deep bidirectional transformers for language understanding}.
\newblock In \emph{Proceedings of the 2019 Conference of the North {A}merican
  Chapter of the Association for Computational Linguistics: Human Language
  Technologies, Volume 1 (Long and Short Papers)}, pages 4171--4186,
  Minneapolis, Minnesota. Association for Computational Linguistics.

\bibitem[{Dodge et~al.(2020)Dodge, Ilharco, Schwartz, Farhadi, Hajishirzi, and
  Smith}]{dodge-etal-2020-fine}
Jesse Dodge, Gabriel Ilharco, Roy Schwartz, Ali Farhadi, Hannaneh Hajishirzi,
  and Noah Smith. 2020.
\newblock Fine-tuning pretrained language models: Weight initializations, data
  orders, and early stopping.
\newblock \emph{arXiv preprint arXiv:2002.06305}.

\bibitem[{Ein-Dor et~al.(2020)Ein-Dor, Halfon, Gera, Shnarch, Dankin, Choshen,
  Danilevsky, Aharonov, Katz, and Slonim}]{ein-dor-etal-2020-active}
Liat Ein-Dor, Alon Halfon, Ariel Gera, Eyal Shnarch, Lena Dankin, Leshem
  Choshen, Marina Danilevsky, Ranit Aharonov, Yoav Katz, and Noam Slonim. 2020.
\newblock \href {https://doi.org/10.18653/v1/2020.emnlp-main.638} {{A}ctive
  {L}earning for {BERT}: {A}n {E}mpirical {S}tudy}.
\newblock In \emph{Proceedings of the 2020 Conference on Empirical Methods in
  Natural Language Processing (EMNLP)}, pages 7949--7962, Online. Association
  for Computational Linguistics.

\bibitem[{Elisha and Dekel(2016)}]{elisha-dekel-2016-wavelet}
Oren Elisha and Shai Dekel. 2016.
\newblock \href {http://jmlr.org/papers/v17/15-203.html} {Wavelet
  decompositions of random forests - smoothness analysis, sparse approximation
  and applications}.
\newblock \emph{Journal of Machine Learning Research}, 17(198):1--38.

\bibitem[{Elisha and Dekel(2017)}]{elisha-dekel-2017-function}
Oren Elisha and Shai Dekel. 2017.
\newblock Function space analysis of deep learning representation layers.
\newblock \emph{arXiv preprint arXiv:1710.03263}.

\bibitem[{Gal and Ghahramani(2016)}]{gal-ghahramani-2016-dropout}
Yarin Gal and Zoubin Ghahramani. 2016.
\newblock \href {https://proceedings.mlr.press/v48/gal16.html} {Dropout as a
  bayesian approximation: Representing model uncertainty in deep learning}.
\newblock In \emph{Proceedings of The 33rd International Conference on Machine
  Learning}, volume~48 of \emph{Proceedings of Machine Learning Research},
  pages 1050--1059, New York, New York, USA. PMLR.

\bibitem[{Gissin and Shalev-Shwartz(2019)}]{gissin-shwartz-2019-discriminative}
Daniel Gissin and Shai Shalev-Shwartz. 2019.
\newblock Discriminative active learning.
\newblock \emph{arXiv preprint arXiv:1907.06347}.

\bibitem[{Grie{\ss}haber et~al.(2020)Grie{\ss}haber, Maucher, and
  Vu}]{griesshaber-etal-2020-fine}
Daniel Grie{\ss}haber, Johannes Maucher, and Ngoc~Thang Vu. 2020.
\newblock \href {https://doi.org/10.18653/v1/2020.coling-main.100} {Fine-tuning
  {BERT} for low-resource natural language understanding via active learning}.
\newblock In \emph{Proceedings of the 28th International Conference on
  Computational Linguistics}, pages 1158--1171, Barcelona, Spain (Online).
  International Committee on Computational Linguistics.

\bibitem[{Gururangan et~al.(2020)Gururangan, Marasovi{\'c}, Swayamdipta, Lo,
  Beltagy, Downey, and Smith}]{gururangan-etal-2020-dont}
Suchin Gururangan, Ana Marasovi{\'c}, Swabha Swayamdipta, Kyle Lo, Iz~Beltagy,
  Doug Downey, and Noah~A. Smith. 2020.
\newblock \href {https://doi.org/10.18653/v1/2020.acl-main.740} {Don{'}t stop
  pretraining: Adapt language models to domains and tasks}.
\newblock In \emph{Proceedings of the 58th Annual Meeting of the Association
  for Computational Linguistics}, pages 8342--8360, Online. Association for
  Computational Linguistics.

\bibitem[{Howard and Ruder(2018)}]{howard-ruder-2018-universal}
Jeremy Howard and Sebastian Ruder. 2018.
\newblock \href {https://doi.org/10.18653/v1/P18-1031} {Universal language
  model fine-tuning for text classification}.
\newblock In \emph{Proceedings of the 56th Annual Meeting of the Association
  for Computational Linguistics (Volume 1: Long Papers)}, pages 328--339,
  Melbourne, Australia. Association for Computational Linguistics.

\bibitem[{Huang et~al.(2016)Huang, Child, Rao, Liu, Satheesh, and
  Coates}]{huang-etal-2016-active}
Jiaji Huang, Rewon Child, Vinay Rao, Hairong Liu, Sanjeev Satheesh, and Adam
  Coates. 2016.
\newblock Active learning for speech recognition: the power of gradients.
\newblock \emph{arXiv preprint arXiv:1612.03226}.

\bibitem[{Ikhwantri et~al.(2018)Ikhwantri, Louvan, Kurniawan, Abisena, Rachman,
  Wicaksono, and Mahendra}]{ikhwantri-etal-2018-multi}
Fariz Ikhwantri, Samuel Louvan, Kemal Kurniawan, Bagas Abisena, Valdi Rachman,
  Alfan~Farizki Wicaksono, and Rahmad Mahendra. 2018.
\newblock \href {https://doi.org/10.18653/v1/W18-3406} {Multi-task active
  learning for neural semantic role labeling on low resource conversational
  corpus}.
\newblock In \emph{Proceedings of the Workshop on Deep Learning Approaches for
  Low-Resource {NLP}}, pages 43--50, Melbourne. Association for Computational
  Linguistics.

\bibitem[{Imaizumi and Fukumizu(2019)}]{imaizumi-fukumizu-2019-deep}
Masaaki Imaizumi and Kenji Fukumizu. 2019.
\newblock Deep neural networks learn non-smooth functions effectively.
\newblock In \emph{AISTATS}.

\bibitem[{Ishibashi and Hino(2021)}]{ishibashi-hideitsu-2021-stopping}
Hideaki Ishibashi and Hideitsu Hino. 2021.
\newblock Stopping criterion for active learning based on error stability.
\newblock \emph{arXiv preprint arXiv:2104.01836}.

\bibitem[{Karamcheti et~al.(2021)Karamcheti, Krishna, Fei-Fei, and
  Manning}]{karamcheti-etal-2021-mind}
Siddharth Karamcheti, Ranjay Krishna, Li~Fei-Fei, and Christopher Manning.
  2021.
\newblock \href {https://doi.org/10.18653/v1/2021.acl-long.564} {Mind your
  outliers! investigating the negative impact of outliers on active learning
  for visual question answering}.
\newblock In \emph{Proceedings of the 59th Annual Meeting of the Association
  for Computational Linguistics and the 11th International Joint Conference on
  Natural Language Processing (Volume 1: Long Papers)}, pages 7265--7281,
  Online. Association for Computational Linguistics.

\bibitem[{Kasai et~al.(2019)Kasai, Qian, Gurajada, Li, and
  Popa}]{kasai-etal-2019-low}
Jungo Kasai, Kun Qian, Sairam Gurajada, Yunyao Li, and Lucian Popa. 2019.
\newblock \href {https://doi.org/10.18653/v1/P19-1586} {Low-resource deep
  entity resolution with transfer and active learning}.
\newblock In \emph{Proceedings of the 57th Annual Meeting of the Association
  for Computational Linguistics}, pages 5851--5861, Florence, Italy.
  Association for Computational Linguistics.

\bibitem[{Lewis and Gale(1994)}]{lewis-gale-1994-sequential}
David~D Lewis and William~A Gale. 1994.
\newblock A sequential algorithm for training text classifiers.
\newblock In \emph{SIGIR’94}, pages 3--12. Springer.

\bibitem[{Li and Roth(2002)}]{li-roth-2002-learning}
Xin Li and Dan Roth. 2002.
\newblock \href {https://aclanthology.org/C02-1150} {Learning question
  classifiers}.
\newblock In \emph{{COLING} 2002: The 19th International Conference on
  Computational Linguistics}.

\bibitem[{Lowell et~al.(2019)Lowell, Lipton, and
  Wallace}]{lowell-etal-2019-practical}
David Lowell, Zachary~C. Lipton, and Byron~C. Wallace. 2019.
\newblock \href {https://doi.org/10.18653/v1/D19-1003} {Practical obstacles to
  deploying active learning}.
\newblock In \emph{Proceedings of the 2019 Conference on Empirical Methods in
  Natural Language Processing and the 9th International Joint Conference on
  Natural Language Processing (EMNLP-IJCNLP)}, pages 21--30, Hong Kong, China.
  Association for Computational Linguistics.

\bibitem[{Margatina et~al.(2022)Margatina, Barrault, and
  Aletras}]{margatina-etal-2022-importance}
Katerina Margatina, Loic Barrault, and Nikolaos Aletras. 2022.
\newblock \href {https://doi.org/10.18653/v1/2022.acl-short.93} {On the
  importance of effectively adapting pretrained language models for active
  learning}.
\newblock In \emph{Proceedings of the 60th Annual Meeting of the Association
  for Computational Linguistics (Volume 2: Short Papers)}, pages 825--836,
  Dublin, Ireland. Association for Computational Linguistics.

\bibitem[{Margatina et~al.(2021)Margatina, Vernikos, Barrault, and
  Aletras}]{margatina-etal-2021-active}
Katerina Margatina, Giorgos Vernikos, Lo{\"\i}c Barrault, and Nikolaos Aletras.
  2021.
\newblock \href {https://doi.org/10.18653/v1/2021.emnlp-main.51} {Active
  learning by acquiring contrastive examples}.
\newblock In \emph{Proceedings of the 2021 Conference on Empirical Methods in
  Natural Language Processing}, pages 650--663, Online and Punta Cana,
  Dominican Republic. Association for Computational Linguistics.

\bibitem[{Mosbach et~al.(2021)Mosbach, Andriushchenko, and
  Klakow}]{mosbach-etal-2021-stability}
Marius Mosbach, Maksym Andriushchenko, and Dietrich Klakow. 2021.
\newblock \href {https://openreview.net/forum?id=nzpLWnVAyah} {On the stability
  of fine-tuning {BERT}: Misconceptions, explanations, and strong baselines}.
\newblock In \emph{International Conference on Learning Representations}.

\bibitem[{Pang and Lee(2004)}]{pang-lee-2004-sentimental}
Bo~Pang and Lillian Lee. 2004.
\newblock \href {https://doi.org/10.3115/1218955.1218990} {A sentimental
  education: Sentiment analysis using subjectivity summarization based on
  minimum cuts}.
\newblock In \emph{Proceedings of the 42nd Annual Meeting of the Association
  for Computational Linguistics ({ACL}-04)}, pages 271--278, Barcelona, Spain.

\bibitem[{Petersen and
  Voigtl{\"a}nder(2018)}]{petersen-voigtlander-2018-optimal}
Philipp~Christian Petersen and Felix Voigtl{\"a}nder. 2018.
\newblock Optimal approximation of piecewise smooth functions using deep relu
  neural networks.
\newblock \emph{Neural networks : the official journal of the International
  Neural Network Society}, 108:296--330.

\bibitem[{Prabhu et~al.(2019)Prabhu, Dognin, and
  Singh}]{prabhu-etal-2019-sampling}
Ameya Prabhu, Charles Dognin, and Maneesh Singh. 2019.
\newblock \href {https://doi.org/10.18653/v1/D19-1417} {Sampling bias in deep
  active classification: An empirical study}.
\newblock In \emph{Proceedings of the 2019 Conference on Empirical Methods in
  Natural Language Processing and the 9th International Joint Conference on
  Natural Language Processing (EMNLP-IJCNLP)}, pages 4058--4068, Hong Kong,
  China. Association for Computational Linguistics.

\bibitem[{Schr{\"o}der et~al.(2022)Schr{\"o}der, Niekler, and
  Potthast}]{schroder-etal-2022-revisiting}
Christopher Schr{\"o}der, Andreas Niekler, and Martin Potthast. 2022.
\newblock \href {https://doi.org/10.18653/v1/2022.findings-acl.172} {Revisiting
  uncertainty-based query strategies for active learning with transformers}.
\newblock In \emph{Findings of the Association for Computational Linguistics:
  ACL 2022}, pages 2194--2203, Dublin, Ireland. Association for Computational
  Linguistics.

\bibitem[{Sener and Savarese(2018)}]{sener-savarese-2018-active}
Ozan Sener and Silvio Savarese. 2018.
\newblock \href {https://openreview.net/forum?id=H1aIuk-RW} {Active learning
  for convolutional neural networks: A core-set approach}.
\newblock In \emph{International Conference on Learning Representations}.

\bibitem[{Settles(2009)}]{settles-2009-active}
Burr Settles. 2009.
\newblock \href
  {http://axon.cs.byu.edu/~martinez/classes/778/Papers/settles.activelearning.pdf}
  {Active learning literature survey}.
\newblock Computer sciences technical report.

\bibitem[{Shelmanov et~al.(2021)Shelmanov, Puzyrev, Kupriyanova, Belyakov,
  Larionov, Khromov, Kozlova, Artemova, Dylov, and
  Panchenko}]{shelmanov-etal-2021-active}
Artem Shelmanov, Dmitri Puzyrev, Lyubov Kupriyanova, Denis Belyakov, Daniil
  Larionov, Nikita Khromov, Olga Kozlova, Ekaterina Artemova, Dmitry~V. Dylov,
  and Alexander Panchenko. 2021.
\newblock \href {https://doi.org/10.18653/v1/2021.eacl-main.145} {Active
  learning for sequence tagging with deep pre-trained models and {B}ayesian
  uncertainty estimates}.
\newblock In \emph{Proceedings of the 16th Conference of the European Chapter
  of the Association for Computational Linguistics: Main Volume}, pages
  1698--1712, Online. Association for Computational Linguistics.

\bibitem[{Siddhant and Lipton(2018)}]{siddhant-lipton-2018-deep}
Aditya Siddhant and Zachary~C. Lipton. 2018.
\newblock \href {https://doi.org/10.18653/v1/D18-1318} {Deep {B}ayesian active
  learning for natural language processing: Results of a large-scale empirical
  study}.
\newblock In \emph{Proceedings of the 2018 Conference on Empirical Methods in
  Natural Language Processing}, pages 2904--2909, Brussels, Belgium.
  Association for Computational Linguistics.

\bibitem[{Srivastava et~al.(2014)Srivastava, Hinton, Krizhevsky, Sutskever, and
  Salakhutdinov}]{srivastava-etal-dropout}
Nitish Srivastava, Geoffrey Hinton, Alex Krizhevsky, Ilya Sutskever, and Ruslan
  Salakhutdinov. 2014.
\newblock \href {http://jmlr.org/papers/v15/srivastava14a.html} {Dropout: A
  simple way to prevent neural networks from overfitting}.
\newblock \emph{Journal of Machine Learning Research}, 15(56):1929--1958.

\bibitem[{Stephenson et~al.(2021)Stephenson, Padhy, Ganesh, Hui, Tang, and
  Chung}]{stephenson-etal-2021-geometry}
Cory Stephenson, Suchismita Padhy, Abhinav Ganesh, Yue Hui, Hanlin Tang, and
  SueYeon Chung. 2021.
\newblock \href {https://openreview.net/forum?id=V8jrrnwGbuc} {On the geometry
  of generalization and memorization in deep neural networks}.
\newblock In \emph{International Conference on Learning Representations}.

\bibitem[{Suzuki(2019)}]{suzuki-2019-adaptivity}
Taiji Suzuki. 2019.
\newblock \href {https://openreview.net/forum?id=H1ebTsActm} {Adaptivity of
  deep re{LU} network for learning in besov and mixed smooth besov spaces:
  optimal rate and curse of dimensionality}.
\newblock In \emph{International Conference on Learning Representations}.

\bibitem[{Suzuki and Nitanda(2021)}]{suzuki-atsushi-2021-deep}
Taiji Suzuki and Atsushi Nitanda. 2021.
\newblock \href
  {https://proceedings.neurips.cc/paper/2021/file/1dacb10f0623c67cb7dbb37587d8b38a-Paper.pdf}
  {Deep learning is adaptive to intrinsic dimensionality of model smoothness in
  anisotropic {B}esov space}.
\newblock In \emph{Advances in Neural Information Processing Systems},
  volume~34, pages 3609--3621. Curran Associates, Inc.

\bibitem[{Vaswani et~al.(2017)Vaswani, Shazeer, Parmar, Uszkoreit, Jones,
  Gomez, Kaiser, and Polosukhin}]{vaswani-etal-2017-attention}
Ashish Vaswani, Noam Shazeer, Niki Parmar, Jakob Uszkoreit, Llion Jones,
  Aidan~N Gomez, {\L}ukasz Kaiser, and Illia Polosukhin. 2017.
\newblock Attention is all you need.
\newblock \emph{Advances in neural information processing systems}, 30.

\bibitem[{Vlachos(2008)}]{vlachos-2008-stopping}
Andreas Vlachos. 2008.
\newblock A stopping criterion for active learning.
\newblock \emph{Computer Speech \& Language}, 22(3):295--312.

\bibitem[{Wolf et~al.(2020)Wolf, Debut, Sanh, Chaumond, Delangue, Moi, Cistac,
  Rault, Louf, Funtowicz, Davison, Shleifer, von Platen, Ma, Jernite, Plu, Xu,
  Le~Scao, Gugger, Drame, Lhoest, and Rush}]{wolf-etal-2020-transformers}
Thomas Wolf, Lysandre Debut, Victor Sanh, Julien Chaumond, Clement Delangue,
  Anthony Moi, Pierric Cistac, Tim Rault, Remi Louf, Morgan Funtowicz, Joe
  Davison, Sam Shleifer, Patrick von Platen, Clara Ma, Yacine Jernite, Julien
  Plu, Canwen Xu, Teven Le~Scao, Sylvain Gugger, Mariama Drame, Quentin Lhoest,
  and Alexander Rush. 2020.
\newblock \href {https://doi.org/10.18653/v1/2020.emnlp-demos.6} {Transformers:
  State-of-the-art natural language processing}.
\newblock In \emph{Proceedings of the 2020 Conference on Empirical Methods in
  Natural Language Processing: System Demonstrations}, pages 38--45, Online.
  Association for Computational Linguistics.

\bibitem[{Yarotsky(2017)}]{yarotsky-2017-error}
Dmitry Yarotsky. 2017.
\newblock Error bounds for approximations with deep relu networks.
\newblock \emph{Neural Networks}, 94:103--114.

\bibitem[{Yu et~al.(2022)Yu, Kong, Zhang, Zhang, and
  Zhang}]{yu-etal-2022-actune}
Yue Yu, Lingkai Kong, Jieyu Zhang, Rongzhi Zhang, and Chao Zhang. 2022.
\newblock \href {https://doi.org/10.18653/v1/2022.naacl-main.102} {{A}c{T}une:
  Uncertainty-based active self-training for active fine-tuning of pretrained
  language models}.
\newblock In \emph{Proceedings of the 2022 Conference of the North American
  Chapter of the Association for Computational Linguistics: Human Language
  Technologies}, pages 1422--1436, Seattle, United States. Association for
  Computational Linguistics.

\bibitem[{Yuan et~al.(2020)Yuan, Lin, and Boyd-Graber}]{yuan-etal-2020-cold}
Michelle Yuan, Hsuan-Tien Lin, and Jordan Boyd-Graber. 2020.
\newblock \href {https://doi.org/10.18653/v1/2020.emnlp-main.637} {Cold-start
  active learning through self-supervised language modeling}.
\newblock In \emph{Proceedings of the 2020 Conference on Empirical Methods in
  Natural Language Processing (EMNLP)}, pages 7935--7948, Online. Association
  for Computational Linguistics.

\bibitem[{Zhang et~al.(2021)Zhang, Wu, Katiyar, Weinberger, and
  Artzi}]{zhang-etal-2021-revisiting}
Tianyi Zhang, Felix Wu, Arzoo Katiyar, Kilian~Q Weinberger, and Yoav Artzi.
  2021.
\newblock \href {https://openreview.net/forum?id=cO1IH43yUF} {Revisiting
  few-sample {BERT} fine-tuning}.
\newblock In \emph{International Conference on Learning Representations}.

\bibitem[{Zhang et~al.(2015)Zhang, Zhao, and LeCun}]{zhang-etal-2015-character}
Xiang Zhang, Junbo Zhao, and Yann LeCun. 2015.
\newblock Character-level convolutional networks for text classification.
\newblock \emph{Advances in neural information processing systems}, 28.

\bibitem[{Zhang et~al.(2017)Zhang, Lease, and Wallace}]{zhang-etal-2017-active}
Ye~Zhang, Matthew Lease, and Byron Wallace. 2017.
\newblock Active discriminative text representation learning.
\newblock In \emph{Proceedings of the AAAI Conference on Artificial
  Intelligence}, volume~31.

\bibitem[{Zhu et~al.(2010)Zhu, Wang, Hovy, and Ma}]{zhu-etal-2010-confidence}
Jingbo Zhu, Huizhen Wang, Eduard Hovy, and Matthew Ma. 2010.
\newblock Confidence-based stopping criteria for active learning for data
  annotation.
\newblock \emph{ACM Transactions on Speech and Language Processing (TSLP)},
  6(3):1--24.

\end{thebibliography}
\bibliographystyle{acl_natbib}

\clearpage
\appendix

\section{Reproducibility}
\label{sec:rep}

\subsection{Dataset statistics}
We report the sizes of the datasets per split in \Cref{tab:dataset-stats}. The datasets contain mainly texts in English.
\begin{table}[t!]
\small
\centering
\begin{tabular}{lrrrr}
\toprule
& \textsc{train} & \textsc{val} & \textsc{test} & \textsc{total} \\
\midrule
\trecb{} & $1,987$ & $159$ & $486$ & $2,632$ \\
\subj{} & $7,000$ & $1,000$ & $2,000$ & $10,000$ \\
\agnb{} & $20,000$ & $2,600$ & $5,000$ & $27,600$ \\
\trec{} & $4,881$ & $452$ & $500$ & $5,833$ \\
\agn{} & $20,000$ & $7,600$ & $7,600$ & $35,200$ \\
\bottomrule
\end{tabular}
\caption{Dataset sizes by splits. Although we do not use a validation set (\textsc{val}) in our experiments, we report its size for completeness. We uniformly subsampled the \agnb{} and \agn{} datasets for shorter computation time.} 
\label{tab:dataset-stats}
\end{table}

\subsection{Models}
We used base and uncased variants of the Transformer models. Specifically, we used ``bert-base-uncased'' for \bert{} and ``google/electra-base-discriminator'' for \electra{}. Both models have $109,514,298$ trainable parameters each.

\subsection{AL methods}
\begin{description}
\item[\textsc{mc}] We use ten inference cycles to approximate the entropy of the output via Monte-Carlo dropout sampling.
\item[\textsc{cs}] We use the [\textsc{cls}] token representation from the Transformer's penultimate layer. We opt for the greedy method described in the original paper \cite{sener-savarese-2018-active}.
\end{description}

\subsection{Preprocessing}
We use the same preprocessing pipeline on all datasets for both \bert{} and \electra{}. We lowercase the tokens, remove non-alphanumeric tokens and truncate the sequence to $200$ tokens.

\subsection{Hyperparameters}
\label{app:hyper}
We used a fixed learning rate of $2 \times 10^{-5}$ for both models. Additionally, we set the gradient clipping to $1$ during training. In the \textsc{st} regime, we trained the model for $5$ epochs and $15$ in \textsc{et}, and \textsc{eta}. For TAPT, we used masked language modeling with $15\%$ of randomly masked tokens and trained the model via self-supervision for $50$ epochs with the learning rate set to $10^{-5}$.

\subsection{Computing infrastructure}
We conducted our experiments on $4 \times$ \textit{AMD Ryzen Threadripper 3970X 32-Core Processors} and $4 \times$ \textit{NVIDIA GeForce RTX 3090} GPUs with $24$GB of RAM. We used \textit{PyTorch} version $1.9.0$ and CUDA $11.4$.

\subsection{Average runtime}

We report the average runtime of experiments in \Cref{tab:runtime}. We ran six sampling methods on five datasets for two models and for five different training regimes. Additionally, we repeated each experiment five times with different seeds ($[1, 2, 3, 4, 5]$). In each experiment, we re-train the model $20$ times on \textsc{trec-2}, \textsc{subj}, and \textsc{agn-2} up to $1,000$ instances ($20$ batches of $50$ instances), and $40$ times on \textsc{trec-6} and \textsc{agn-4} up to $2,000$ instances ($40$ batches of size $50$). In total, we ran $300$ AL experiments.  
\begin{table}
\centering
\small
\begin{tabular}{lcc}
\toprule
& \bert{} & \electra{} \\
\midrule
\trecb{} & $32.4$ & $31.1$ \\
\subj{} & $40.8$ & $39.2$ \\
\agnb{} & $71.4$ & $70.3$ \\
\trec{} & $68.4$ & $67.1$ \\
\agn{} & $82.3$ & $75.7$ \\
\bottomrule
\end{tabular}
\caption{Experiment duration in minutes for both models across datasets. We report the average runtime over five different runs and six different sampling methods (five AL methods and random sampling).}
\label{tab:runtime}
\end{table}

\section{Experiments}

\begin{figure*}
\centering
\begin{subfigure}{0.49\textwidth}
  \centering
  \includegraphics[width=\linewidth]{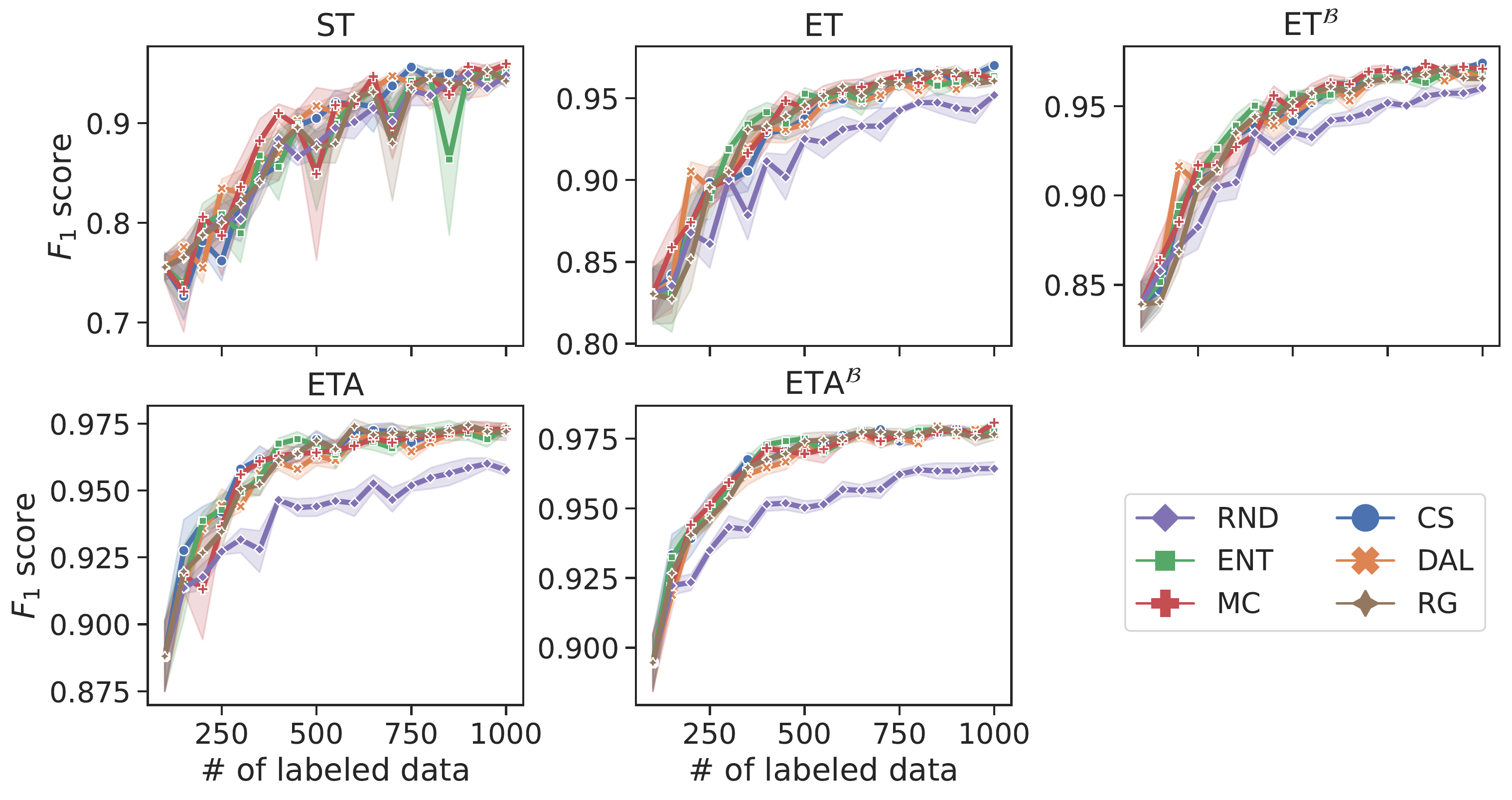}
  \caption{\textsc{trec-2}; \bert{}}
  \label{fig:short}
\end{subfigure}
\begin{subfigure}{0.49\textwidth}
  \centering
  \includegraphics[width=\linewidth]{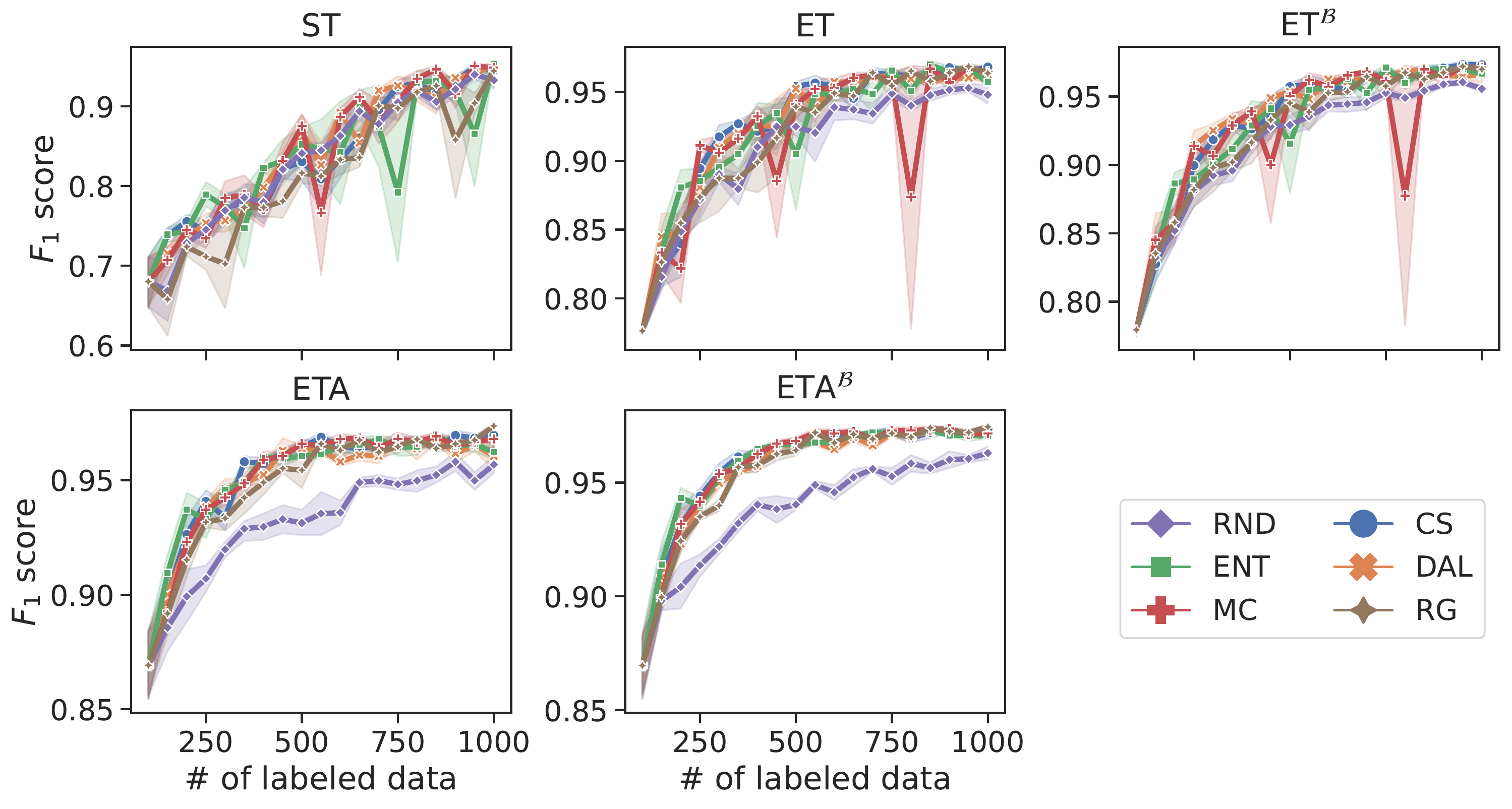}
  \caption{\textsc{trec-2}; \electra{}}
  \label{fig:short}
\end{subfigure}
\begin{subfigure}{0.49\textwidth}
  \centering
  \includegraphics[width=\linewidth]{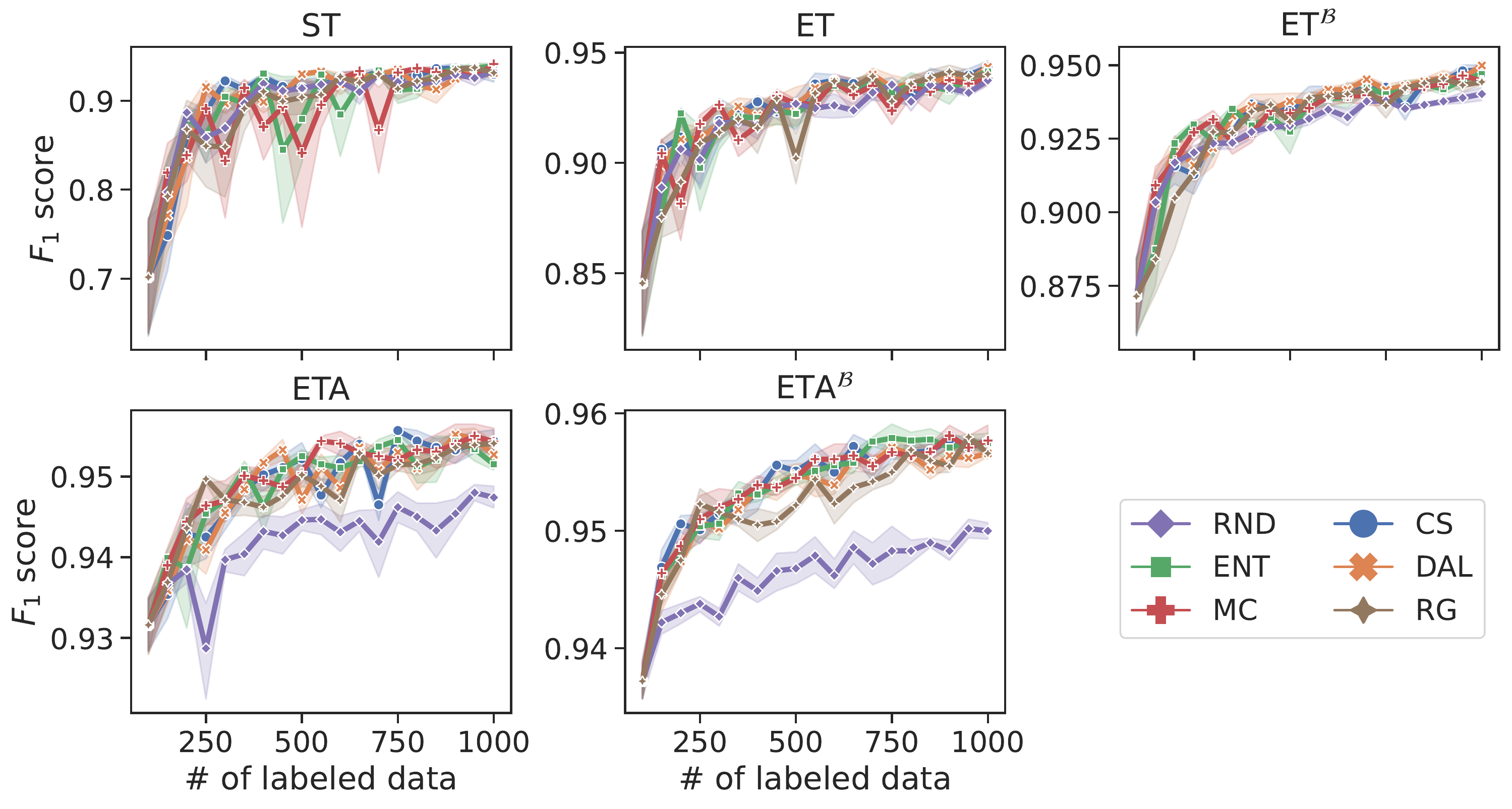}
  \caption{\textsc{subj}; \bert{}}
  \label{fig:short}
\end{subfigure}
\begin{subfigure}{0.49\textwidth}
  \centering
  \includegraphics[width=\linewidth]{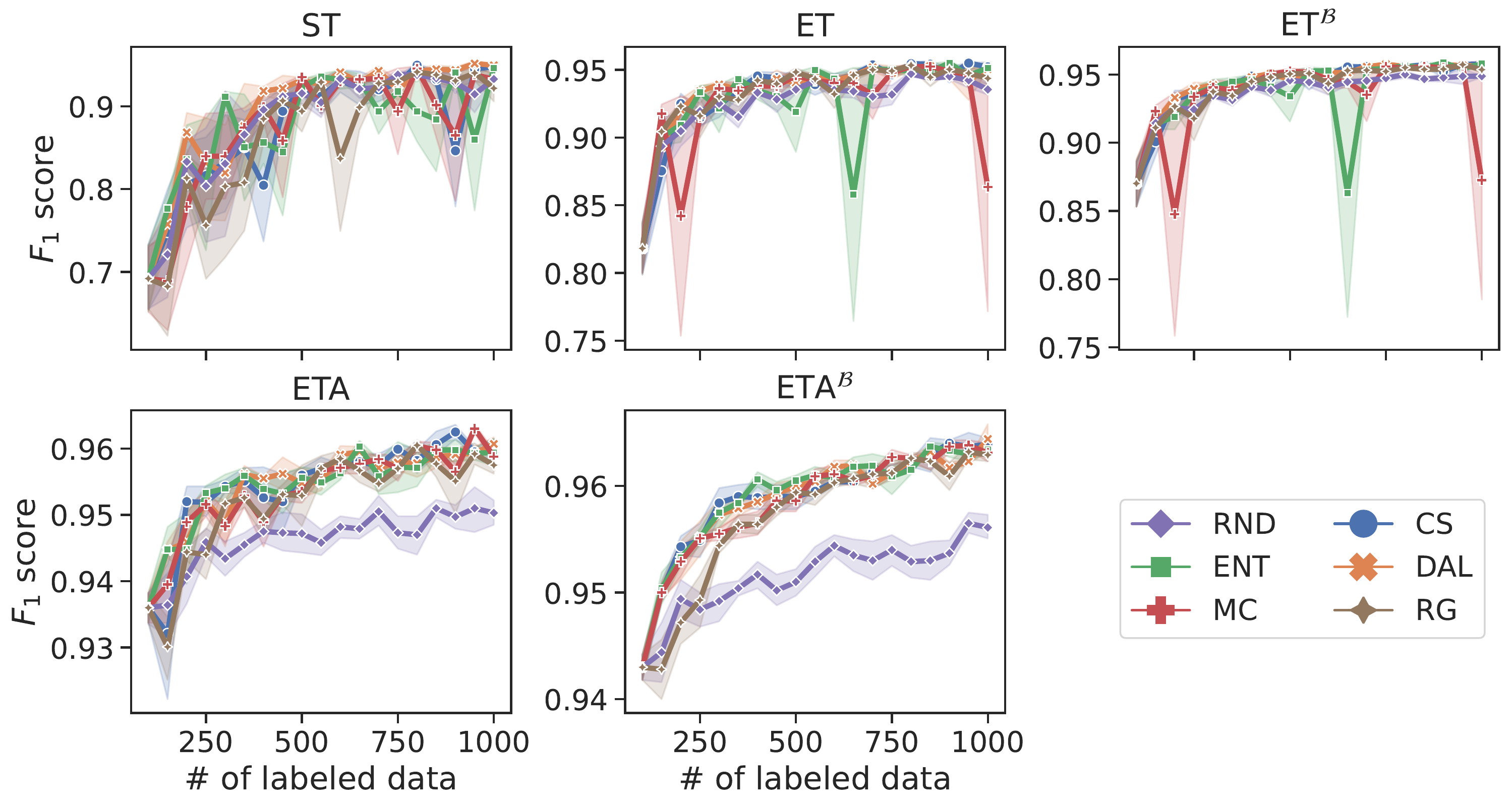}
  \caption{\textsc{subj}; \electra{}}
  \label{fig:short}
\end{subfigure}
\begin{subfigure}{0.49\textwidth}
  \centering
  \includegraphics[width=\linewidth]{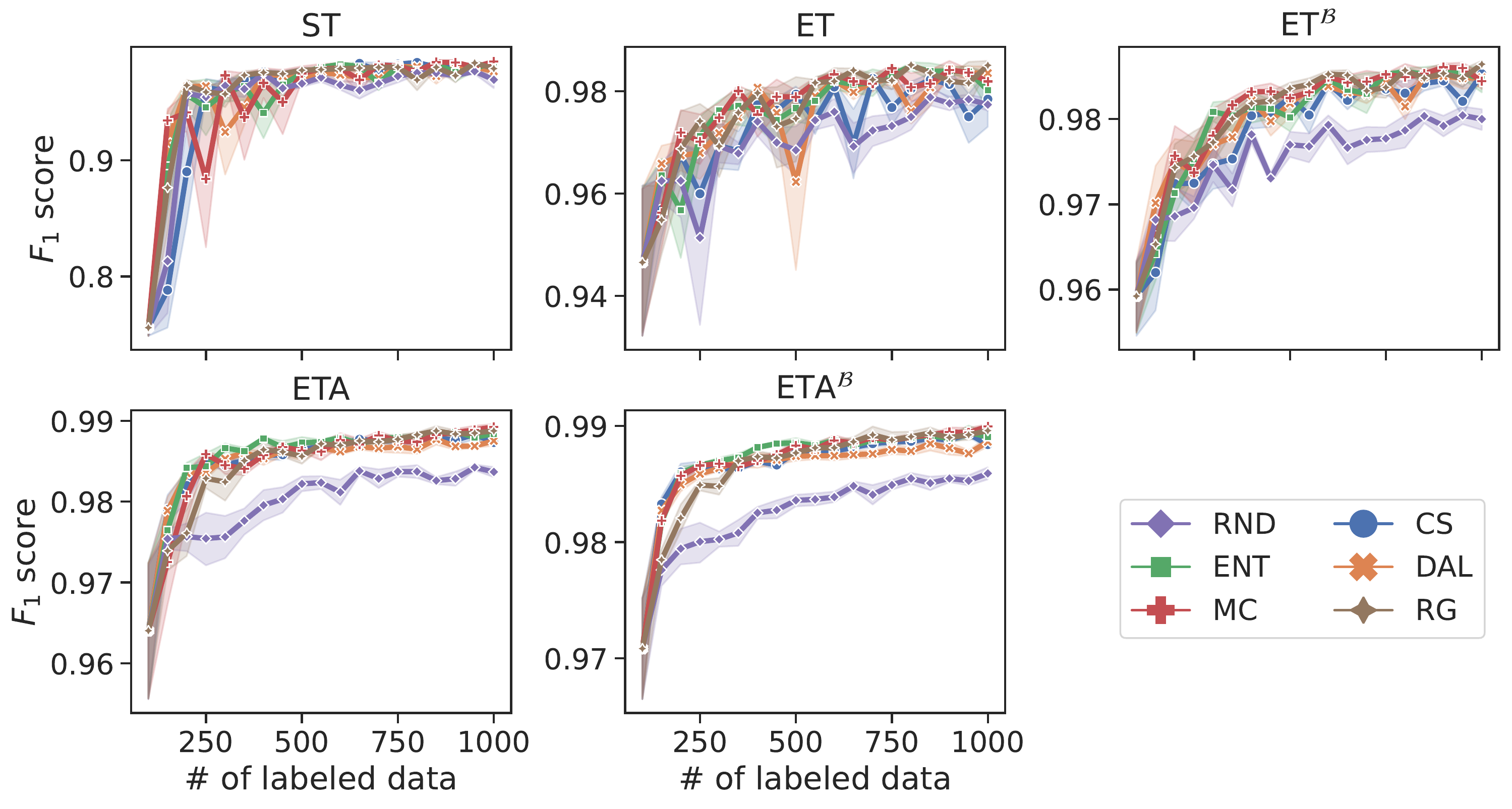}
  \caption{\textsc{agn-2}; \bert{}}
  \label{fig:short}
\end{subfigure}
\begin{subfigure}{0.49\textwidth}
  \centering
  \includegraphics[width=\linewidth]{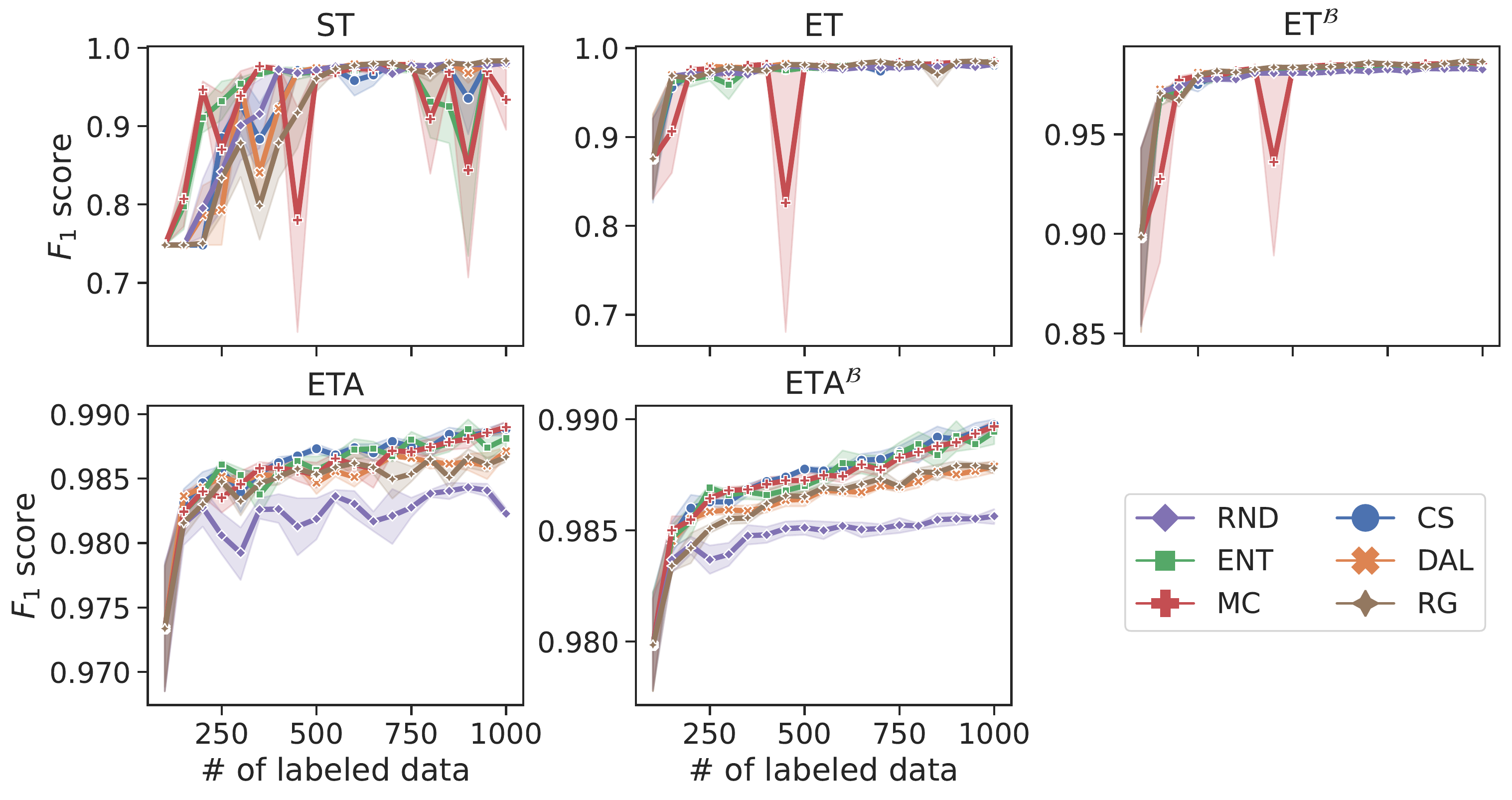}
  \caption{\textsc{agn-2}; \electra{}}
  \label{fig:short}
\end{subfigure}
\begin{subfigure}{0.49\textwidth}
  \centering
  \includegraphics[width=\linewidth]{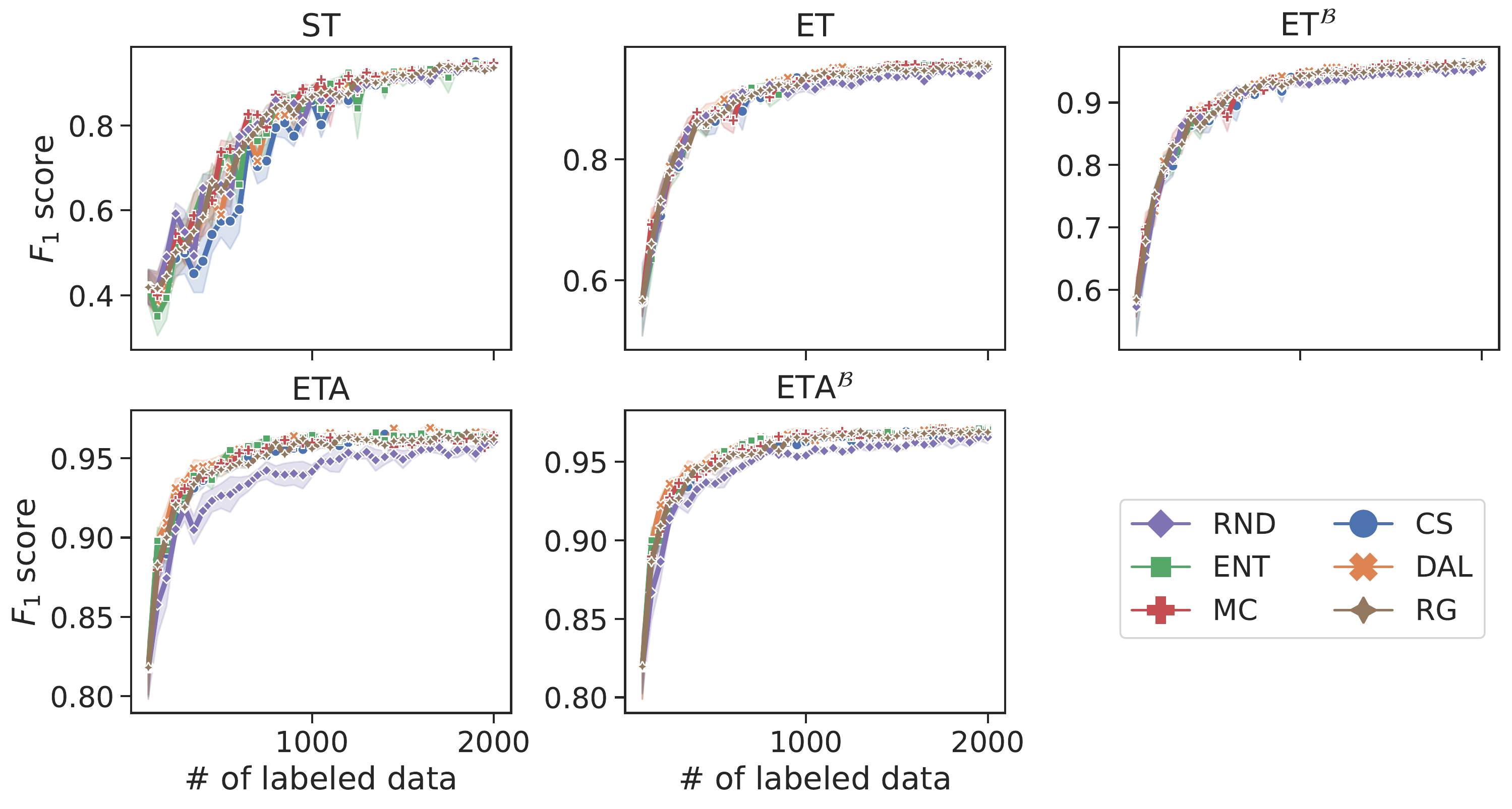}
  \caption{\textsc{trec-6}; \bert{}}
  \label{fig:short}
\end{subfigure}
\begin{subfigure}{0.49\textwidth}
  \centering
  \includegraphics[width=\linewidth]{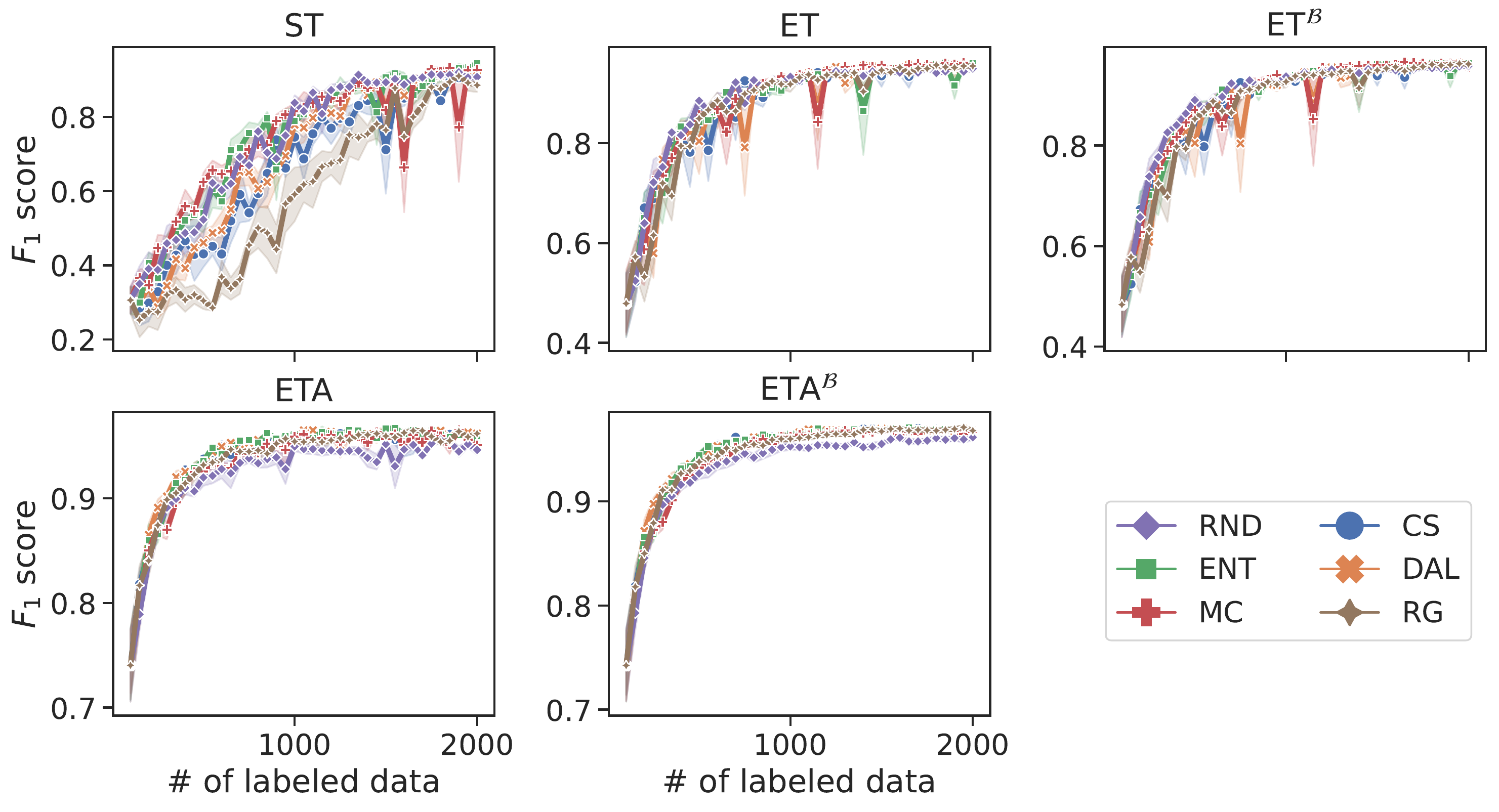}
  \caption{\textsc{trec-6}; \electra{}}
  \label{fig:short}
\end{subfigure}
\begin{subfigure}{0.49\textwidth}
  \centering
  \includegraphics[width=\linewidth]{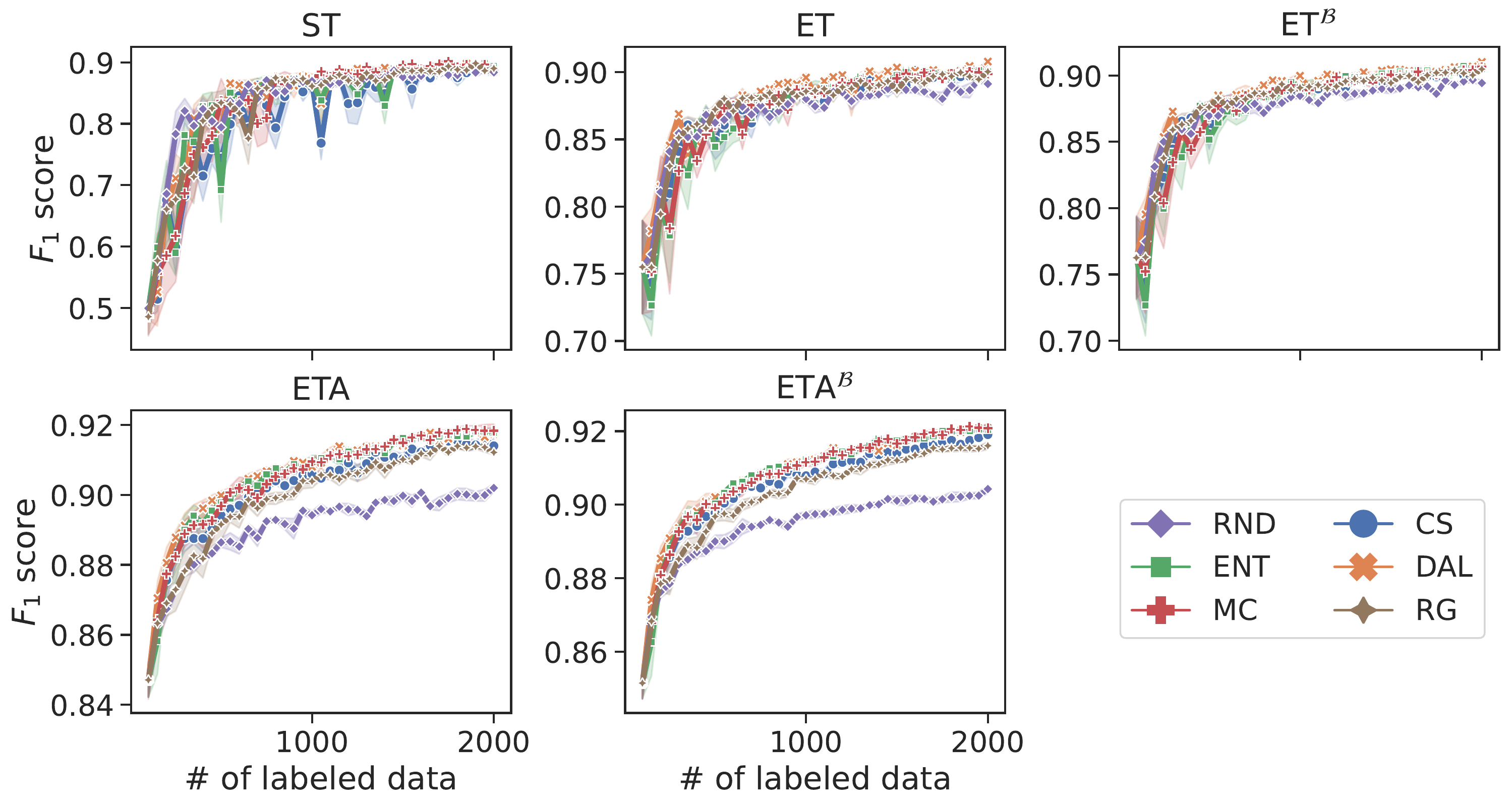}
  \caption{\textsc{agn-4}; \bert{}}
  \label{fig:short}
\end{subfigure}
\begin{subfigure}{0.49\textwidth}
  \centering
  \includegraphics[width=\linewidth]{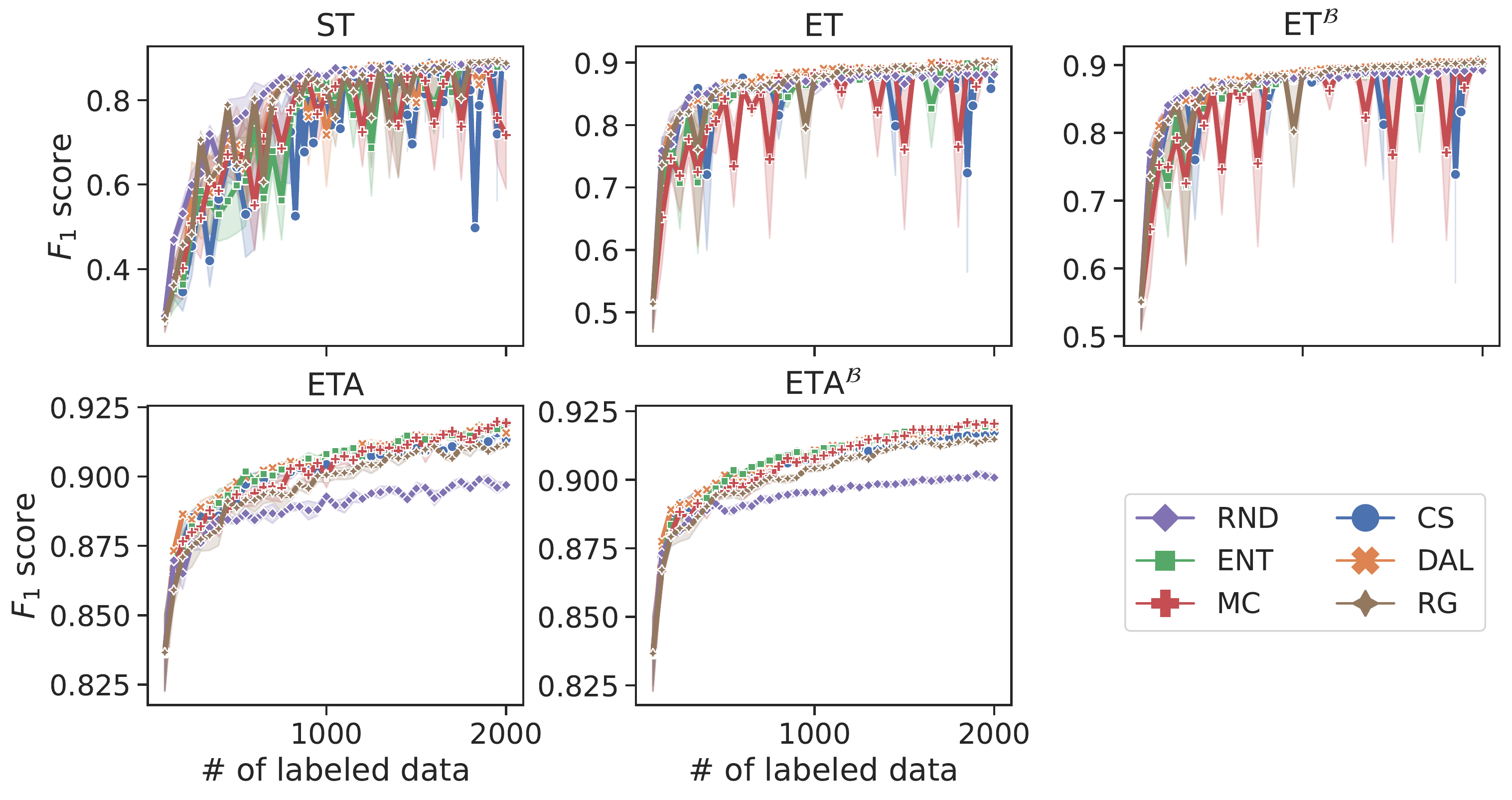}
  \caption{\textsc{agn-4}; \electra{}}
  \label{fig:short}
\end{subfigure}
\caption{AL performance curves for different training regimes across datasets and models. Random sampling (purple rhombs) serves as a baseline. Best viewed on a computer screen.}
\label{fig:app-curves}
\end{figure*}

We report the experiments that were omitted from the main part of the paper due to space constraints. \Cref{fig:app-curves} shows the active learning performance curves across the used datasets and for both models (\bert{} and \electra{}). For the \textsc{eta} and \textsc{eta$^\mathcal{B}$} training regimes, we observe a consistent improvement in performance compared to random sampling. We report the results for \electra{} in \Cref{tab:app-auc}, where we observed similar patterns as with \bert{} (cf.~\Cref{tab:auc} in the main part of the paper). On top of that, our early stopping method reduces the variance of the results compared to other training regimes, as shown in \Cref{tab:std}.

\begin{figure}[t!]
\small
\centering
\begin{subfigure}{\linewidth}
  \centering
  \includegraphics[width=.75\linewidth]{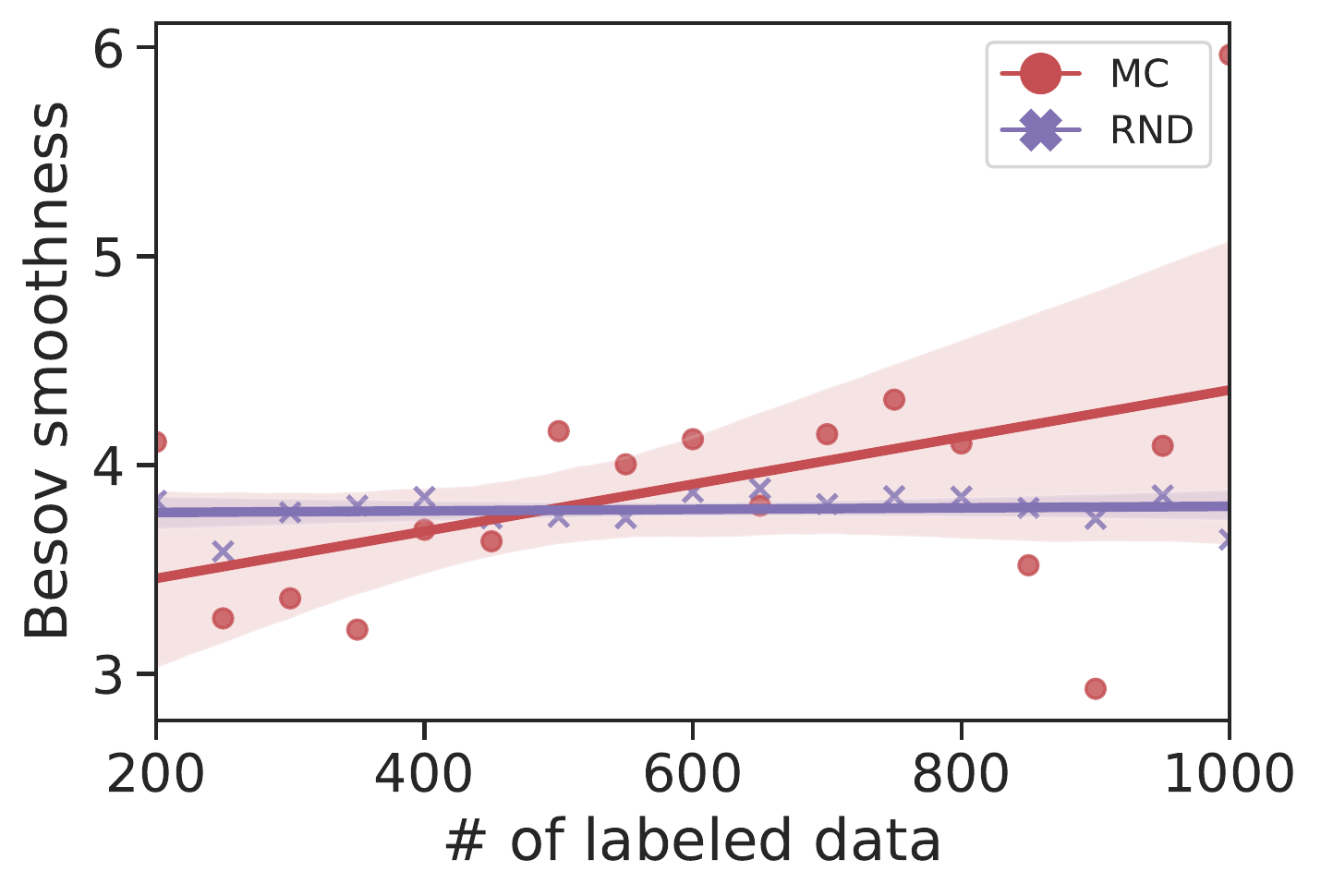}
  \subcaption{\textsc{trec-2}}
  \label{fig:as-trec-2}
\end{subfigure}
\begin{subfigure}{\linewidth}
  \centering
  \includegraphics[width=.75\linewidth]{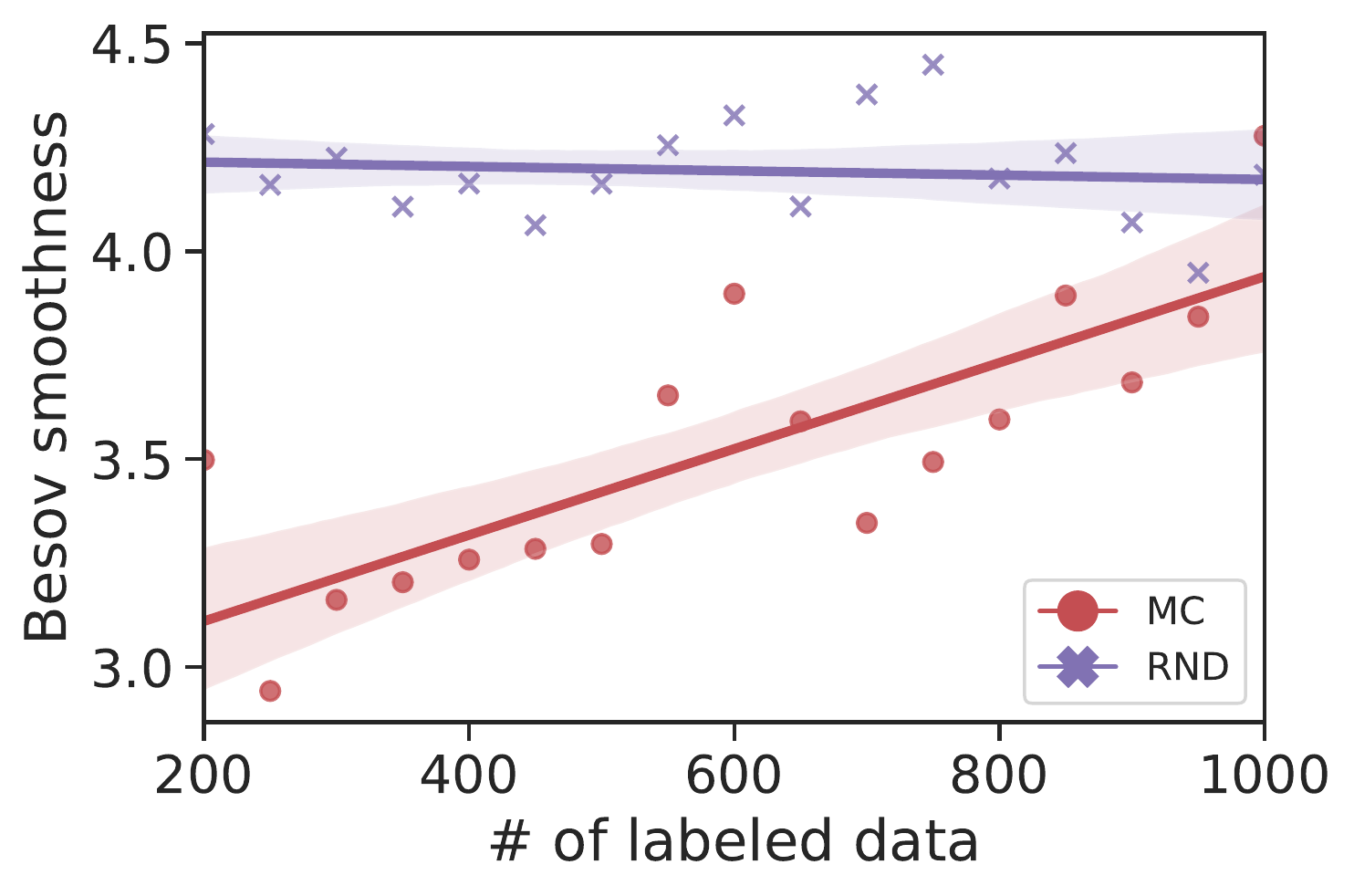}
  \subcaption{\textsc{agn-2}}
  \label{fig:alp-trec-2}
\end{subfigure}

\begin{subfigure}{\linewidth}
  \centering
  \includegraphics[width=.75\linewidth]{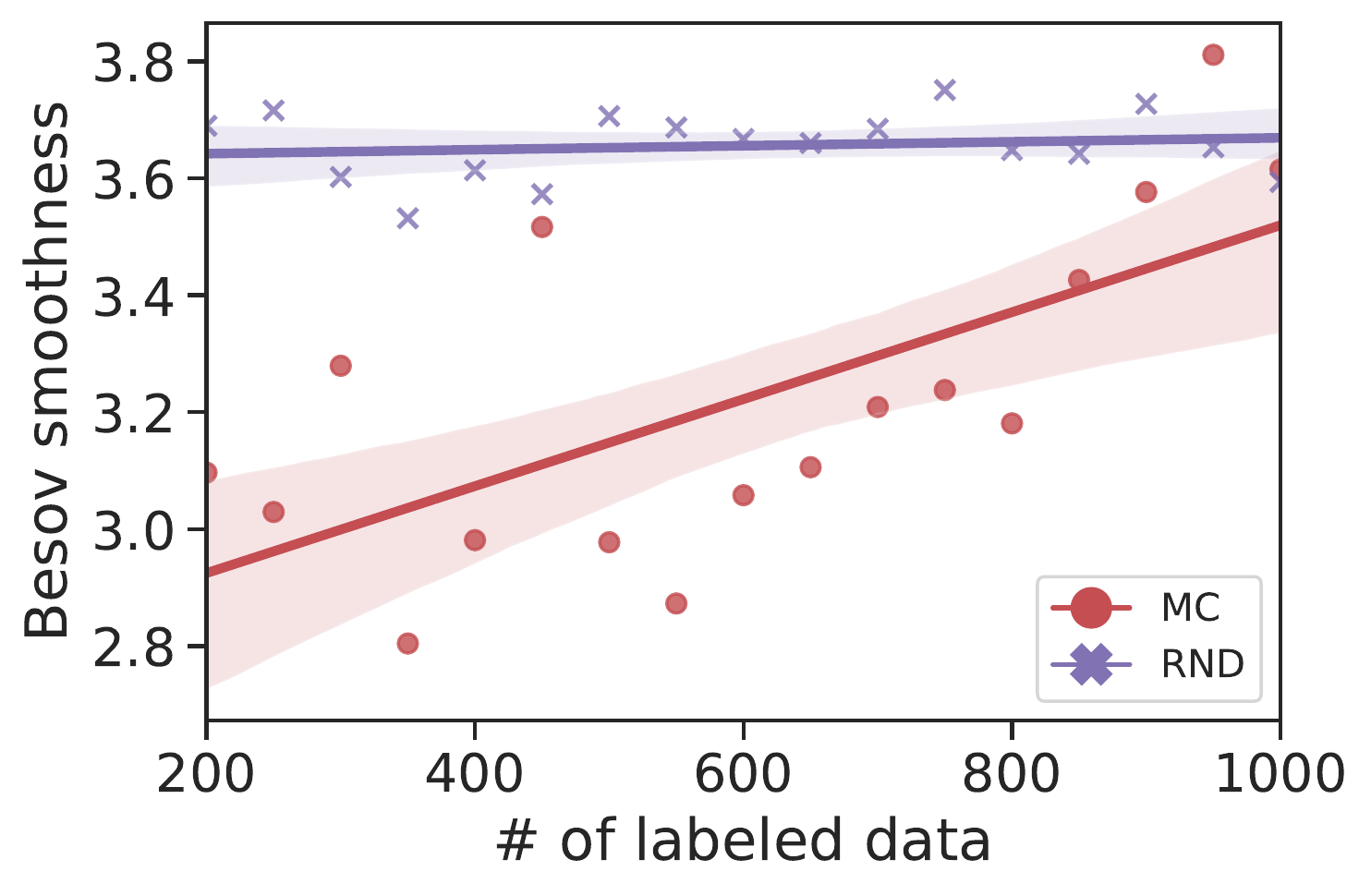}
  \subcaption{\textsc{subj}}
  \label{fig:as-subj}
\end{subfigure}
\begin{subfigure}{\linewidth}
  \centering
  \includegraphics[width=.75\linewidth]{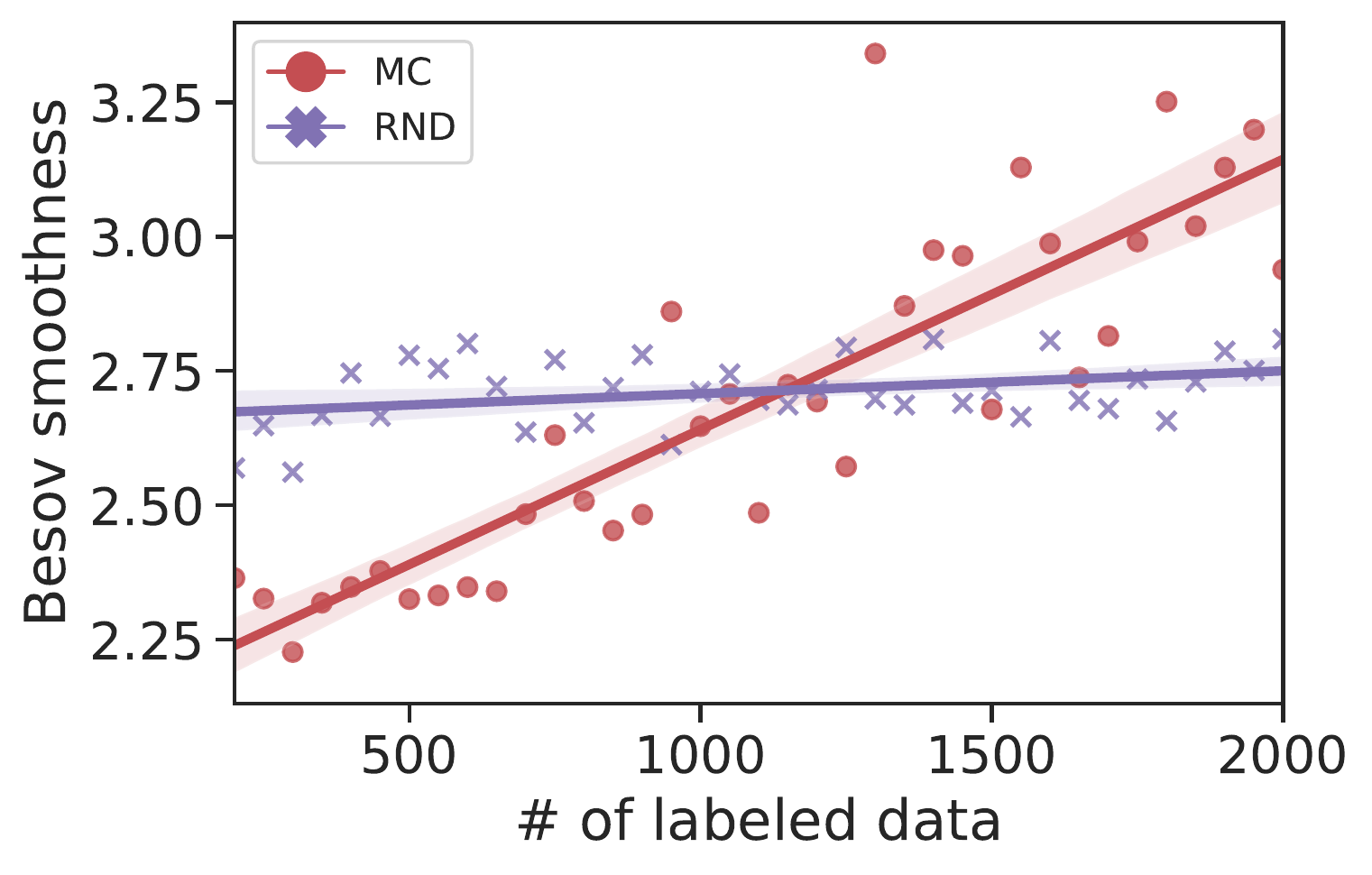}
  \subcaption{\textsc{trec-6}}
  \label{fig:alp-subj}
\end{subfigure}
\begin{subfigure}{\linewidth}
  \centering
  \includegraphics[width=.75\linewidth]{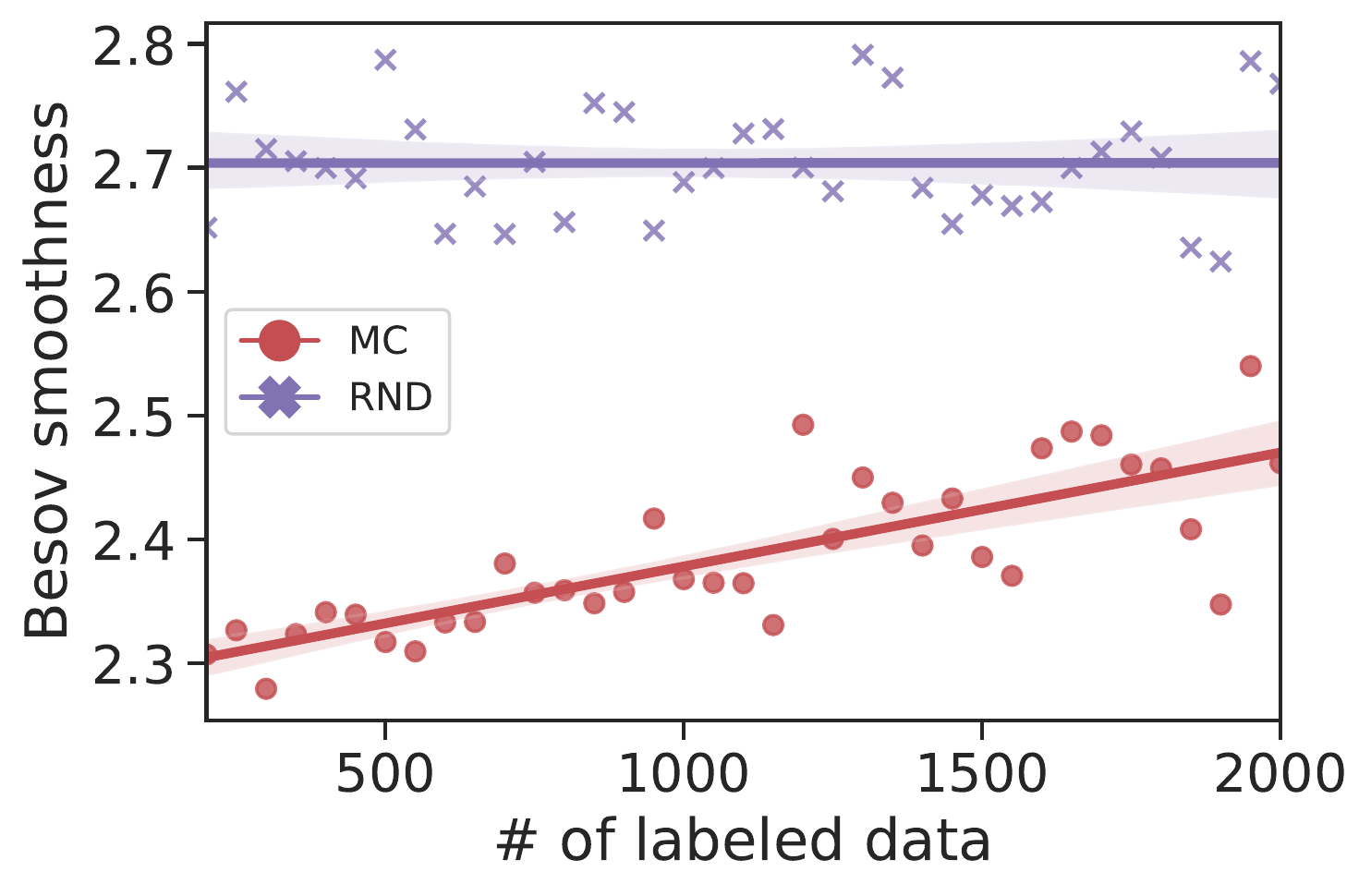}
  \subcaption{\textsc{agn-4}}
  \label{fig:alp-agn-4}
\end{subfigure}
\caption{Besov smoothness of actively acquired samples with \textsc{mc} (red) compared to the smoothness of random samples (purple).}
\label{fig:app-active-sample}
\end{figure}

\Cref{fig:app-active-sample} shows the relationship between the Besov smoothness of random and active samples. We report the smoothness of samples for each dataset. We observe a similar pattern, with a rising smoothness of actively acquired samples.

\begin{table}[t]
\centering
\small
\begin{tabular}{lrrrrrrr}
\toprule
& & \rnd{} & \ent{} & \mc{} & \cs{} & \dal{} & \rg{} \\
\midrule
\multirow{5}{*}{\rotatebox[origin=c]{90}{\textsc{trec-2}}}
& \textsc{st} & $.831$ & $.829$ & $.836$ & $.835$ & $.840$ & $.805$\\
& \textsc{et} & $.910$ & $.924$ & $.918$ & $.928$ & $.927$ & $.919$\\
& \textsc{et\nospacetext{$^\mathcal{B}$}} & $.919$ & $.930$ & $.926$ & $.934$ & $.936$ & $.927$\\
& \textsc{eta} & $.932$ & $.953$ & $.953$ & $.953$ & $.951$ & $.949$\\
& \textsc{eta\nospacetext{$^\mathcal{B}$}} & $.939$ & $.959$ & $.958$ & $.959$ & $.956$ & $.956$\\
\midrule
\multirow{5}{*}{\rotatebox[origin=c]{90}{\textsc{subj}}}
& \textsc{st} & $.880$ & $.872$ & $.870$ & $.871$ & $.898$ & $.860$\\
& \textsc{et} & $.926$ & $.927$ & $.925$ & $.935$ & $.936$ & $.935$\\
& \textsc{et\nospacetext{$^\mathcal{B}$}} & $.938$ & $.937$ & $.934$ & $.944$ & $.946$ & $.942$\\
& \textsc{eta} & $.946$ & $.955$ & $.954$ & $.955$ & $.955$ & $.952$\\
& \textsc{eta\nospacetext{$^\mathcal{B}$}} & $.952$ & $.959$ & $.959$ & $.959$ & $.959$ & $.957$\\
\midrule
\multirow{5}{*}{\rotatebox[origin=c]{90}{\textsc{agn-2}}}
& \textsc{st} & $.867$ & $.901$ & $.891$ & $.850$ & $.850$ & $.823$\\
& \textsc{et} & $.963$ & $.963$ & $.954$ & $.963$ & $.966$ & $.965$\\
& \textsc{et\nospacetext{$^\mathcal{B}$}} & $.969$ & $.971$ & $.962$ & $.971$ & $.971$ & $.972$\\
& \textsc{eta} & $.977$ & $.981$ & $.981$ & $.982$ & $.981$ & $.980$\\
& \textsc{eta\nospacetext{$^\mathcal{B}$}} & $.980$ & $.983$ & $.983$ & $.983$ & $.982$ & $.982$\\
\midrule
\multirow{5}{*}{\rotatebox[origin=c]{90}{\textsc{trec-6}}}
& \textsc{st} & $.604$ & $.645$ & $.636$ & $.549$ & $.561$ & $.461$\\
& \textsc{et} & $.837$ & $.848$ & $.839$ & $.817$ & $.811$ & $.814$\\
& \textsc{et\nospacetext{$^\mathcal{B}$}} & $.843$ & $.858$ & $.847$ & $.821$ & $.813$ & $.816$\\
& \textsc{eta} & $.897$ & $.917$ & $.905$ & $.905$ & $.905$ & $.901$\\
& \textsc{eta\nospacetext{$^\mathcal{B}$}} & $.906$ & $.925$ & $.917$ & $.915$ & $.914$ & $.911$\\
\midrule
\multirow{5}{*}{\rotatebox[origin=c]{90}{\textsc{agn-4}}}
& \textsc{st} & $.793$ & $.706$ & $.713$ & $.688$ & $.755$ & $.750$\\
& \textsc{et} & $.857$ & $.844$ & $.824$ & $.845$ & $.866$ & $.857$\\
& \textsc{et\nospacetext{$^\mathcal{B}$}} & $.868$ & $.855$ & $.836$ & $.857$ & $.874$ & $.866$\\
& \textsc{eta} & $.888$ & $.903$ & $.901$ & $.901$ & $.904$ & $.897$\\
& \textsc{eta\nospacetext{$^\mathcal{B}$}} & $.893$ & $.907$ & $.905$ & $.905$ & $.907$ & $.901$\\
\bottomrule
\end{tabular}
\caption{\auc{} for random sampling and different AL methods across datasets and training regimes for \electra{}. The results are averaged over five runs with different seeds.
}
\label{tab:app-auc}
\end{table}
\begin{table}[t]
\small
\centering
\begin{tabular}{lccccc}
\toprule
& \st{} & \et{} & \etar{} & \etb{} & \textsc{eta$^\mathcal{B}$} \\
\midrule
\trecb{} & $.0093$ & $.0053$ & $.0045$ & $.0026$ & $\textbf{.0022}$\\
\subj{} & $.0117$ & $.0045$ & $.0032$ & $.0013$ & $\textbf{.0008}$\\
\agnb{} & $.0100$ & $.0036$ & $.0020$ & $.0009$ & $\textbf{.0005}$\\
\trec{} & $.0134$ & $.0081$ & $.0074$ & $.0032$ & $\textbf{.0027}$\\
\agn{} & $.0118$ & $.0048$ & $.0045$ & $.0022$ & $\textbf{.0014}$\\
\bottomrule
\end{tabular}
\caption{Average standard deviation for different training regimes. The results are averaged across models and AL methods. \textbf{Bold} numbers indicate regimes with the lowest standard deviation for a particular dataset.}
\label{tab:std}
\end{table}

\end{document}